\documentclass[lettersize,journal]{IEEEtran}
\usepackage{amsmath,amsfonts}
\usepackage{algorithmic}
\usepackage{algorithm}
\usepackage{array}
\usepackage[caption=false,font=normalsize,labelfont=sf,textfont=sf]{subfig}
\usepackage{textcomp}
\usepackage{stfloats}
\usepackage{url}
\usepackage{verbatim}
\usepackage{graphicx}
\usepackage{cite}
\usepackage{amsmath,amssymb,amsfonts}
\usepackage[inline]{enumitem}
\usepackage{algorithmic}
\usepackage{graphicx}
\usepackage{textcomp}
\usepackage{booktabs}
\usepackage{multirow}
\usepackage{hyperref}
\hypersetup{
colorlinks=true,linkcolor=red,citecolor=blue,urlcolor=black}

\usepackage[utf8]{inputenc}
\usepackage[T1]{fontenc}
\usepackage{float}
\usepackage{adjustbox}
\usepackage{array}
\usepackage{color}
\usepackage{etoolbox}

\definecolor{unlabeled}{rgb}{0.0, 0.0, 0.0}
\definecolor{car}{rgb}{0.39215686274509803, 0.5882352941176471, 0.9607843137254902}
\definecolor{bicycle}{rgb}{0.39215686274509803, 0.9019607843137255, 0.9607843137254902}
\definecolor{motorcycle}{rgb}{0.11764705882352941, 0.23529411764705882, 0.5882352941176471}
\definecolor{truck}{rgb}{0.3137254901960784, 0.11764705882352941, 0.7058823529411765}
\definecolor{othervehicle}{rgb}{0.0, 0.0, 1.0}
\definecolor{person}{rgb}{1.0, 0.11764705882352941, 0.11764705882352941}
\definecolor{bicyclist}{rgb}{1.0, 0.1568627450980392, 0.7843137254901961}
\definecolor{motorcyclist}{rgb}{0.5882352941176471, 0.11764705882352941, 0.35294117647058826}
\definecolor{road}{rgb}{1.0, 0.0, 1.0}
\definecolor{parking}{rgb}{1.0, 0.5882352941176471, 1.0}
\definecolor{sidewalk}{rgb}{0.29411764705882354, 0.0, 0.29411764705882354}
\definecolor{otherground}{rgb}{0.6862745098039216, 0.0, 0.29411764705882354}
\definecolor{building}{rgb}{1.0, 0.7843137254901961, 0.0}
\definecolor{fence}{rgb}{1.0, 0.47058823529411764, 0.19607843137254902}
\definecolor{vegetation}{rgb}{0.0, 0.6862745098039216, 0.0}
\definecolor{trunk}{rgb}{0.5294117647058824, 0.23529411764705882, 0.0}
\definecolor{terrain}{rgb}{0.5882352941176471, 0.9411764705882353, 0.3137254901960784}
\definecolor{pole}{rgb}{1.0, 0.9411764705882353, 0.5882352941176471}
\definecolor{trafficsign}{rgb}{1.0, 0.0, 0.0}

\definecolor{unlabeled_n}{rgb}{0.0, 0.0, 0.0}
\definecolor{barrier_n}{rgb}{0.4392156863, 0.5019607843, 0.5647058823529412}
\definecolor{bicycle_n}{rgb}{0.8627, 0.0784, 0.2353}
\definecolor{bus_n}{rgb}{1.0000, 0.4980, 0.3137}
\definecolor{car_n}{rgb}{1.0000, 0.6196, 0.0000}
\definecolor{c_v_n}{rgb}{0.9137, 0.5882, 0.2745}
\definecolor{motorcycle_n}{rgb}{1.0000, 0.2392, 0.3882}
\definecolor{pedestrian_n}{rgb}{0.0000, 0.0000, 0.9020}
\definecolor{cone_n}{rgb}{0.1843, 0.3098, 0.3098}
\definecolor{trailer_n}{rgb}{1.0000, 0.5490, 0.0000}
\definecolor{truck_n}{rgb}{1.0000, 0.3882, 0.2784}
\definecolor{drive_n}{rgb}{0.0000, 0.8118, 0.7490}
\definecolor{other_flat_n}{rgb}{0.6863, 0.0000, 0.2941}
\definecolor{sidewalk_n}{rgb}{0.2941, 0.0000, 0.2941}
\definecolor{terrain_n}{rgb}{0.4392, 0.7059, 0.2353}
\definecolor{manmade_n}{rgb}{0.8706, 0.7216, 0.5294}
\definecolor{vegetation_n}{rgb}{0.0000, 0.6863, 0.0000}

\newcommand\semcolor[1][black]{\textcolor{#1}{\rule{2.2mm}{2.2mm}}}

\usepackage[capitalise]{cleveref} 

\begin{document}

\title{MASS: Multi-Attentional Semantic Segmentation of LiDAR Data for Dense Top-View Understanding}

\author{Kunyu Peng$^{1}$, Juncong Fei$^{2,3}$, Kailun Yang$^{1}$, Alina Roitberg$^{1}$, Jiaming Zhang$^{1}$, Frank Bieder$^{2}$,\\Philipp Heidenreich$^{3}$, Christoph Stiller$^{2}$, and Rainer Stiefelhagen$^{1}$
\thanks{This work was funded by the German Federal Ministry for Economic Affairs and Energy within the project ``Methoden und Maßnahmen zur Absicherung von KI basierten Wahrnehmungsfunktionen f\"ur das automatisierte Fahren (KI-Absicherung)''. 
This work was also supported in part by the Federal Ministry of Labor and Social Affairs (BMAS) through the AccessibleMaps project under Grant 01KM151112, in part by the University of Excellence through the ``KIT Future Fields'' project, and in part by Hangzhou SurImage Company Ltd.
The authors would like to thank the consortium for the successful cooperation.
\textit{(Corresponding author: Juncong Fei.)}}
\thanks{$^{1}$Authors are with Institute for Anthropomatics and Robotics, Karlsruhe Institute of Technology, Germany (e-mail: \{kunyu.peng, kailun.yang, alina.roitberg, jiaming.zhang, rainer.stiefelhagen\}@kit.edu).}
\thanks{$^{2}$Authors are with Institute for Measurement and Control Systems, Karlsruhe Institute of Technology, Germany (e-mail: juncong.fei@partner.kit.edu, frank.bieder@kit.edu, stiller@kit.edu).}
\thanks{$^{3}$Authors are with Stellantis, Opel Automobile GmbH, Germany.}
\thanks{Code will be made publicly available at github.com/KPeng9510/MASS}
}

\maketitle

\begin{abstract}
At the heart of all automated driving systems is the ability to sense the surroundings, \textit{e.g.,} through semantic segmentation of LiDAR sequences, which experienced a remarkable progress due to the release of large datasets such as SemanticKITTI and nuScenes-LidarSeg. While most previous works focus on \emph{sparse} segmentation of the LiDAR input, \emph{dense} output masks provide self-driving cars with almost complete environment information. In this paper, we introduce MASS - a Multi-Attentional Semantic Segmentation model specifically built for dense top-view understanding of the driving scenes. Our framework operates on pillar- and occupancy features and comprises three attention-based building blocks: (1) a keypoint-driven graph attention, (2) an LSTM-based attention computed from a vector embedding of the spatial input, and (3) a pillar-based attention, resulting in a dense $360^\circ$ segmentation mask. With extensive experiments on both, SemanticKITTI and nuScenes-LidarSeg, we quantitatively demonstrate the effectiveness of our model, outperforming the state of the art by $19.0\%$ on SemanticKITTI and reaching $30.4\%$ in mIoU on nuScenes-LidarSeg, where MASS is the first work addressing the dense segmentation task. Furthermore, our multi-attention model is shown to be very effective for 3D object detection validated on the KITTI-3D dataset, showcasing its high generalizability to other tasks related to 3D vision.
\end{abstract}
 

\IEEEpeerreviewmaketitle

\section{Introduction}
\IEEEPARstart{A}{}reliable semantic understanding of the surroundings is crucial for automated driving.
To this end, multi-modal input captured, \textit{e.g.}, by cameras, LiDARs, and radars is frequently leveraged in automated vehicles~\cite{camera_lidar_integration,radar_camera,issafe}.
Semantic segmentation is one of the most essential tasks in automated driving systems since it predicts pixel- or point-level labels for the surrounding environment according to different input modalities.
Over the past few years, semantic segmentation employing 2D Convolutional Neural Networks (CNNs) has evolved to a well developed field, where FCN~\cite{fcn}, DeepLab~\cite{deeplabv2}, and ERFNet~\cite{erfnet,romera2019bridging} represent prominent architectures.
Recent emergence of large-scale datasets for semantic segmentation of 3D data, such as SemanticKITTI~\cite{semantic_kitti} and nuScenes-LidarSeg~\cite{nuscenes} has allowed the community to go beyond the conventional 2D semantic segmentation and develop novel methods operating on 3D LiDAR point clouds~\cite{gao2020we}.

\begin{figure}[t]
\begin{center}
\includegraphics[width=1.0\columnwidth]{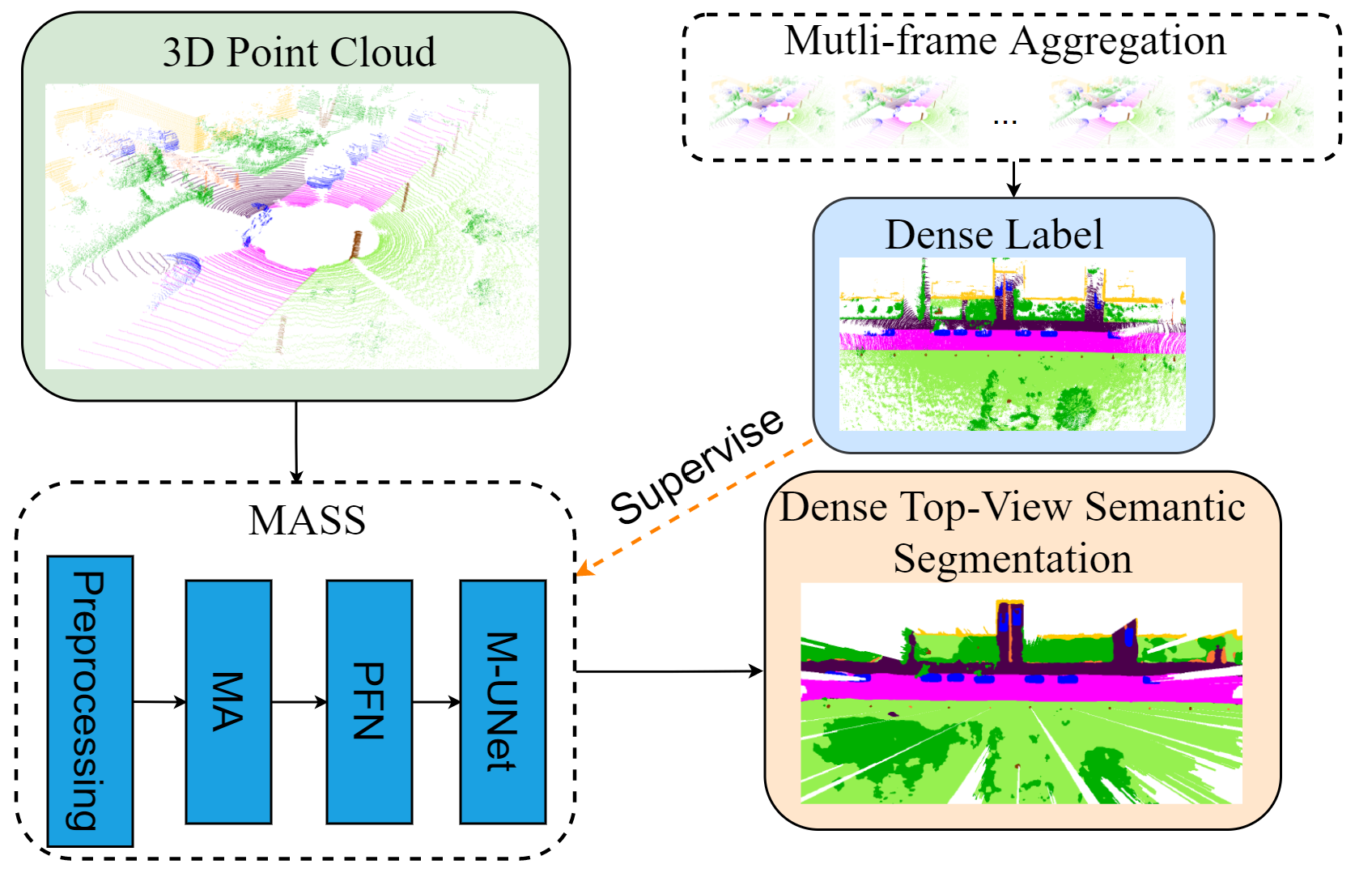}
\end{center}
\vskip-3ex
\caption{An overview of the dense top-view semantic segmentation based on the proposed MASS framework, where LiDAR data is painted by its semantic label on the top left. In the model structure, MA denotes our multi-attention mechanism, PFN denotes the pillar feature net, and M-UNet denotes the modified UNet. The network is supervised by the labeled grid cell and evaluated by the visible region shown by the occupancy map.
}
\label{fig:fig1}
\end{figure}

3D point cloud data generated through LiDAR sensors has multiple advantages over 2D data~\cite{lidar_survey}.
Such point cloud data complements traditional 2D image projection techniques and has direct access to the depth information, leading to a richer spatial information about the surrounding environment.
Furthermore, 3D LiDAR point clouds directly incorporate distance and direction information, while camera-based systems can only infer through generated images to reconstruct distance- and orientation-related information.
Of course, LiDAR data also brings certain challenges.
Since 3D point cloud data is sparse, unordered, and irregular in terms of its spatial shape, it is not straightforward to transfer mature 2D CNN-based approaches to LiDAR data.
To solve this problem, PointNet~\cite{pointnet} extracts point-level features, whereas PointPillars~\cite{pointpillars} forms a top-view pseudo image based on high-dimensional pillar-level features in order to utilize a 2D backbone for 3D object detection.
The pillar feature net is also leveraged in our PillarSegNet architecture, which is put forward as the backbone in our framework.
Some works focus on predicting point-level semantic class for each LiDAR point given a 3D point cloud such as the approaches proposed by~\cite{rangenet++,scan_based,multi_scale_interaction,cheng20212}, which realize sparse segmentation.
In contrast to these approaches, our PillarSegNet generates dense top-view semantic segmentation given a sparse 3D point cloud as the input, which can even accurately yield predictions on those locations without any LiDAR measurements (see Fig.~\ref{fig:fig1}).
This dense interpretation is clearly beneficial to essential upper-level operating functions such as the top view based navigation for automated driving~\cite{multimodal_av}.

In this paper, we introduce a Multi-Attentional Semantic Segmentation (MASS) framework, which aggregates local- and global features, and thereby boosts the performance of dense top-view semantic segmentation.
Top-view semantic segmentation map generation is challenging and often requires multi-stage processing, as such frameworks need to implicitly solve a multitude of sub-tasks, such as ground plane estimation, 3D object detection, route planning, road segmentation (see~\cite{predicting_semantic_map} for further details).
Compared with sparse 3D LiDAR point semantic segmentation~\cite{gndnet}, our 2D dense top-view semantic segmentation harvests richer environment information which may offer more useful cues to these related tasks as aforementioned.
Precisely, MASS is composed of Multi-Attention (MA) mechanisms, a pillar feature net (PFN), and a modified UNet (M-UNet) utilized for dense top-view semantic segmentation, as depicted in Fig.~\ref{fig:fig1}.
Our MA mechanisms comprise three attention-based building blocks: (1) a keypoint-driven graph attention, (2) an LSTM-based attention computed from a vector embedding of the spatial input, and (3) a pillar-based attention.
The proposed MASS model is first evaluated on the SemanticKITTI dataset~\cite{semantic_kitti} to verify its performance compared with the state-of-the-art surround-view prediction work~\cite{bieder2020exploiting} and then validated on the nuScenes-LidarSeg dataset~\cite{nuscenes}, where our framework is the first addressing the dense semantic segmentation task.
Finally,  we validate the effectiveness of PointPillars enhancement with our MA mechanism in terms of cross-task generalization.

This work is an extension of our conference paper~\cite{fei2021pillarsegnet}, which has been extended with the novel MA mechanism design, a detailed description of the proposed PillarSegNet backbone model, along with an extended set of experiments on multiple datasets.
In summary, the main contributions are:
\begin{itemize}
    \item We introduce MASS, a Multi-Attentional Semantic Segmentation framework for dense top-view surrounding understanding. We present an end-to-end method PillarSegNet to approach dense semantic grid map estimation as the backbone of our MASS framework, by using only sparse single-sweep LiDAR data.
    \item We propose Multi-Attention (MA) mechanisms composed of two novel attentions and pillar attention to better aggregate features from different perspectives and to boost the performance of dense top-view semantic segmentation given 3D point cloud input.
    \item Experiments and qualitative comparisons are conducted firstly on SemanticKITTI~\cite{semantic_kitti}, nuScenes-LidarSeg~\cite{nuscenes}, and then on the KITTI-3D dataset~\cite{kitti}, to verify the effectiveness of MA separately for dense top-view semantic segmentation and 3D object detection.
    \item A comprehensive analysis is presented on dense top-view semantic surrounding understanding with different attention setups individually on SemanticKITTI, nuScenes-LidarSeg, and KITTI-3D datasets.
\end{itemize}

\section{Related Works}

\subsection{Image Semantic Segmentation and Attention Mechanism}

Dense pixel-wise semantic segmentation has been largely driven by the development of natural datasets~\cite{kitti,deep_multi_modal} and architectural advances since the pioneering Fully Convolutional Networks (FCNs)~\cite{fcn} and early encoder-decoder models~\cite{unet,segnet}.
Extensive efforts have been made to enrich and enlarge receptive fields with context aggregation sub-module designs like dilated convolutions~\cite{multi_scale} and pyramid pooling~\cite{deeplabv2,pspnet}.
In the Intelligent Transportation Systems (ITS) field, real-time segmentation architectures~\cite{erfnet,omnirange} and surrounding-view perception platforms~\cite{restricted,ooss} are constructed for efficient and complete semantic scene understanding.

Another cluster of works takes advantage of the recent self-attention mechanism in transformers~\cite{attention} to harvest long-range contextual information by adaptively weighing features either in the temporal~\cite{attention} or in the spatial~\cite{omnirange,danet} domain.
With focus set on scene segmentation, DANet~\cite{danet} integrates channel- and position attention modules to model associations between any pair of channels or pixels.
In ViT~\cite{vision_transformer} and SETR~\cite{setr}, transformer is directly applied to sequences of image patches for recognition and segmentation tasks.
In Attention Guided LSTM~\cite{spatio_temporal_lstm}, a visual attention model is used to dynamically pool the convolutional features to capture the most important locations, both spatially and temporally.
In Graph Attention Convolution~\cite{graph_attention_convolution}, the kernels are carved into specific shapes for structured feature learning, selectively focusing on the relevant neighboring nodes.
FeaStNet~\cite{feastnet}, sharing a similar spirit, learns to establish correlations between filter weights and graph neighborhoods with arbitrary connectivity.
Concurrent attention design has also been exploited to learn more discriminative features~\cite{omnirange,danet,tanet}.
For example, TANet~\cite{tanet} collectively considers channel-, point-, and voxel-wise attention by stacking them to aggregate multi-level highlighted features. 

While self-attention mechanism has been widely applied in image-based scene parsing, it is underresearched in the field of semantic segmentation of LiDAR input.
We leverage such attention operations to better aggregate features from different points of view and propose a generic multi-attentional framework for dense semantic segmentation with improved discriminative representations.

\begin{figure*}[!t]
\begin{center}
\includegraphics[width=2.0\columnwidth]{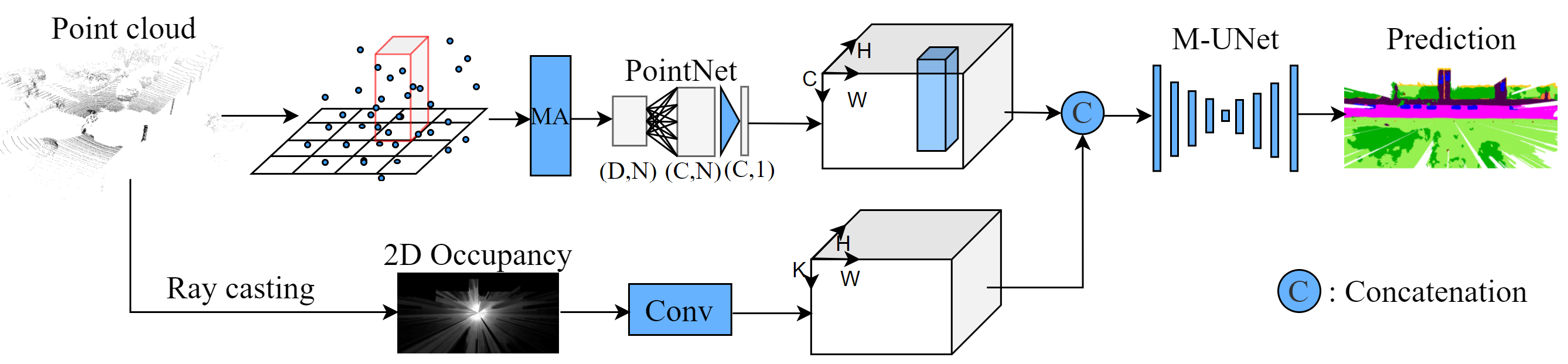}
\end{center}
\caption{Overview of the proposed \mbox{MASS} framework.
Given a 3D point cloud obtained from LiDAR, \mbox{MASS} first executes \emph{pillar-level feature} encoding and computes optional \emph{2D occupancy features} in two parallel streams. 
The point cloud is first rasterized into several pillars and MA generates attention values for these pillars.
The attended pillar-level features are extracted through the PointNet~\cite{pointnet} architecture,
whereas the observability features are encoded from a 2D occupancy map generated through ray casting. Both features will be combined through the concatenation operation.
Then, we leverage a \mbox{modified UNet} to predict a dense top-view semantic grid map from the aggregated features. The final depicted prediction result is filtered by the 2D occupancy map to exclude the occluded areas.
}
\label{fig:pipeline}
\end{figure*}

\subsection{LiDAR Point Cloud Semantic Segmentation}
Unlike image-based scene parsing, the interest in LiDAR point cloud semantic segmentation has been rapidly blossoming until very recently with the appearance of large-scale datasets~\cite{semantic_kitti,nuscenes,semanticposs,semi_lidar}, which provide rich data for supervised training and open up the application in $360^\circ$ point-wise surrounding understanding.  
Since the introduction of PointNet~\cite{pointnet}, many learning-based methods have emerged.
The SqueezeSeg family~\cite{squeezeseg,squeezesegv2} projects the 3D point cloud into 2D pseudo images for processing, and plenty of subsequent methods follow this trend by mapping the 3D LiDAR data under a forward-facing view or a bird's eye view, and thereby inherit the advancements in image semantic segmentation using 2D fully convolutional networks.
RangeNet++~\cite{rangenet++} exploits a transformation to obtain spherical images and employs 2D convolutions for semantic segmentation.
The SalsaNet family~\cite{salsanet,salsanext} presents fast architectures, which have been validated either in the top-down bird's eye view~\cite{salsanet} or in the spherical range view (\textit{i.e.}, panoramic view)~\cite{salsanext}.  
Triess~\textit{et al.}~\cite{scan_based} leverage a scan unfolding and a cyclic padding mechanism to recover the context information at the horizontal panorama borders, which helps to eliminate point occlusions during the spherical projection in~\cite{rangenet++}.
Such unfolding and ring padding are similar to those in panoramic scene parsing~\cite{pass}, and thus we consider that this line of research can benefit from the latest progress in omnidirectional image segmentation like attention mechanisms~\cite{omnirange}.

Instead of using range images, some methods utilize a grid-based representation to perform top-view semantic segmentation~\cite{gndnet,bieder2020exploiting,semantic_grid_estimation,monocular_semantic_occupancy_grid_mapping,sparse_dense}. 
GndNet~\cite{gndnet} uses PointNet~\cite{pointnet} to extract point-wise features and semantically segment ground sparse data.
PolarNet~\cite{polarnet} quantizes the points into grids using their polar bird's eye view coordinates. 
In a recent work, Bieder~\textit{et al.}~\cite{bieder2020exploiting} transform 3D LiDAR data into a multi-layer grid map representation to enable an efficient dense top-view semantic segmentation of LiDAR data.
However, it comes with information loss when generating the grid maps and thus performs unsatisfactorily on small-scale objects.
To address these issues, we put forward a novel end-to-end method termed PillarSegNet, first appeared in our conference work~\cite{fei2021pillarsegnet}, which directly learns features from the point cloud and thereby mitigates the potential information loss. 
PillarSegNet divides the single-sweep LiDAR point cloud into a set of pillars, and generates a dense semantic grid map using such sparse LiDAR data.
Further, the proposed MASS framework intertwines PillarSegNet and multiple attention mechanisms to boost the segmentation performance.

There are additional methods that directly operate on 3D LiDAR data to infer per-point semantics using 3D learning schemes~\cite{pass3d,randlanet,cylindrical_asymmetrical} and various point cloud segmentation-based ITS applications~\cite{multiscale_pointwise,low_channel_lidar,backpack_lidar,fast_pointcloud_ground}.
Moreover, LiDAR data segmentation is promising to be fused with image-based panoramic scene parsing towards a complete geometric and semantic surrounding understanding~\cite{camera_lidar_integration,ooss,semanticvoxels}.

\section{MASS: Proposed Framework}
In this section, we introduce MASS - a new framework for Multi-Attentional Semantic Segmentation given LiDAR point cloud data as input.
First, we put forward a backbone model for dense top-view semantic segmentation given single sweep LiDAR data as input.
Then, we utilize Multi-Attention (MA) mechanisms to aggregate local- and global features, and guide the network to specifically focus on feature map regions which are decisive for our task.

Conceptually, MASS comprises two building blocks: PillarSegNet -- a novel dense top-view semantic segmentation architecture, which extracts pillar-level features in an end-to-end fashion, and an MA mechanism, with an overview provided in Fig.~\ref{fig:pipeline}.
The proposed MA mechanism itself covers three attention-based techniques: a key-node based graph attention, an LSTM attention with dimensionality reduction of the spatial embedding, and a pillar attention derived from the voxel attention in TANet~\cite{tanet}. 
In the following, key principles of PillarSegNet and the proposed MA mechanisms are detailed.

\subsection{PillarSegNet Model}
A central component of our framework is PillarSegNet -- a novel model for dense top-view semantic segmentation of sparse single LiDAR sweep input.
In contrast to the previously proposed grid-map based method~\cite{bieder2020exploiting}, PillarSegNet directly constructs pillar-level features in an end-to-end fashion and then predicts dense top-view semantic segmentation.
In addition to the pillar-level feature, occupancy feature is also utilized in the PillarSegNet model as aforementioned to aggregate additional free-space information generated through an optional feature branch, which is verified to be critical for improving dense top-view semantic segmentation performance compared with the model only utilizing pillar feature.

PillarSegNet comprises a pillar feature net derived from PointPillars~\cite{pointpillars}, an optional occupancy feature encoding branch, a modified UNet architecture as the 2D backbone, and a dense semantic segmentation head realized by a logits layer.
In later sections, the extensive experiments will verify that leveraging pillar feature net from~\cite{pointpillars} generates better representation than the grid-map-based state-of-the-art method~\cite{bieder2020exploiting}.

\textbf{Pillar Feature Encoding.}
Since 3D point cloud does not have regular shapes compared with 2D images, mature 2D CNN-based approaches cannot directly aggregate point cloud features.
In order to utilize well-established approaches based on 2D convolutions, we first rasterize the 3D point cloud into a set of pillars on the top view, then pillar-level feature is extracted through the pillar feature net and, finally, a pseudo image is formed on the top view.

In the following, $C$ marks the dimensionality of the point encoding before being fed into the pillar feature net, $P$ denotes the maximum number of pillars, and the maximum number of augmented LiDAR points inside a pillar is $N$.
We note that only non-empty pillars are considered.
If the generated pillars or the augmented LiDAR points have not reached the aforementioned maximum numbers, zero padding is leveraged to generate a fixed-size pseudo image.
If the numbers are higher than the desired numbers, random sampling is employed to assure the needed dimensionality.
Consequently, the size of the  tensor passed to PointNet in the next step is therefore $(P, N, C)$. The point feature is encoded through PointNet~\cite{pointnet} composed of fully connected layers sharing weights among points together with \emph{BatchNorm} and \emph{ReLU} layers to extract a high-level representation. Then, pillar-level feature is generated through the \emph{max} operation among all the points inside a pillar and the tensor representation is changed to $(P, C)$.
Finally, these pillars are scattered back according to their coordinates on the $xy$ plane to generate a top-view pseudo image for the input of the modified UNet backbone for semantic segmentation.

\textbf{Occupancy Feature.}
Occupancy feature encodes observability through ray casting simulating the physical generation process of each LiDAR point.
This feature is highly important for dense top-view semantic segmentation as it encodes the critical free-space information.

\begin{figure}
\begin{center}
\includegraphics[width=1.0\columnwidth]{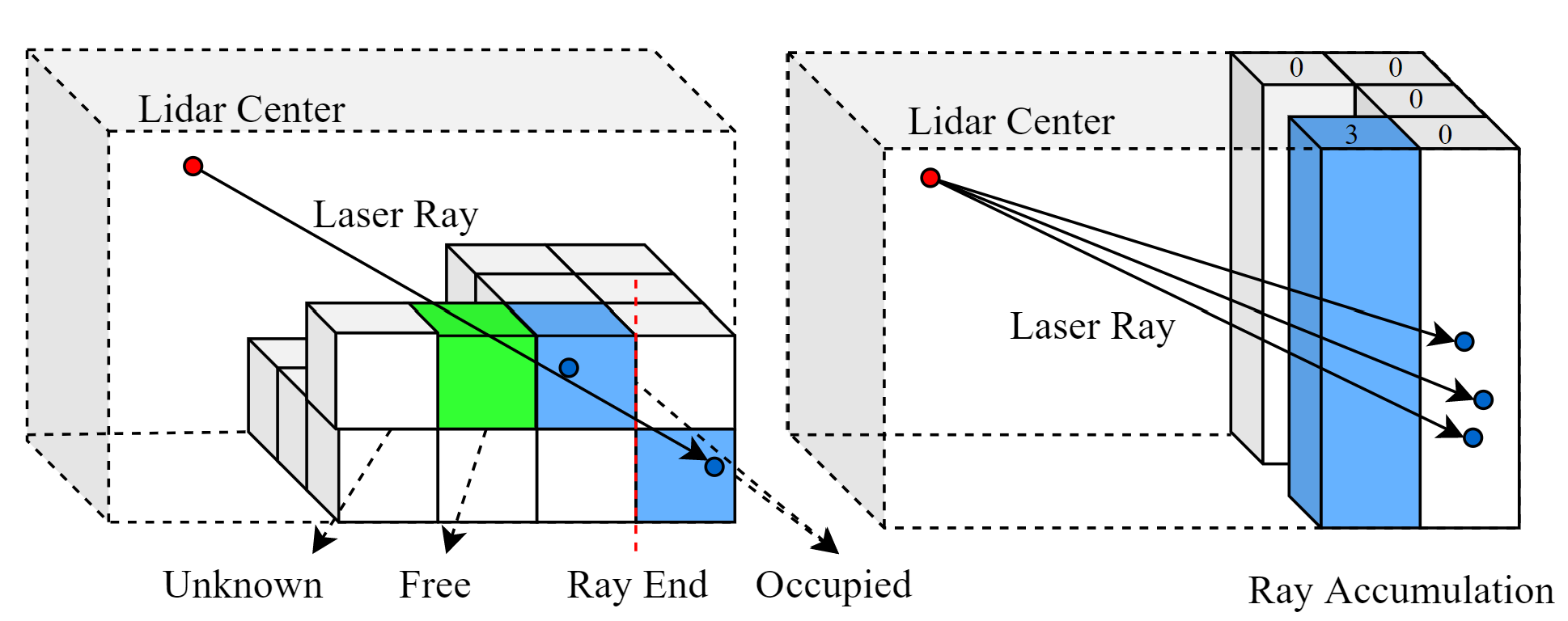}
\end{center}
\vskip-3ex
\caption{A generation procedure comparison between visibility feature (left) and observability feature (right), where the red line on left figure denotes the end of the laser ray.}
\label{fig:generatation_procedure}
\end{figure}

There are two kinds of occupancy encoding approaches: visibility-based and observability-based.
According to the existing work proposed by~\cite{hu20wysiwyg}, \emph{visibility} feature is leveraged to encode 3D sparse occupancy generated based on the 3D point cloud.
The procedure of ray casting approach to generate visibility feature is depicted in Fig.~\ref{fig:generatation_procedure}.
The point cloud is firstly rasterized as 3D grids and has the same spatial resolution on the top-view with the pseudo image for a better fusion.
The initial states of all grid cells are set as unknown. For each LiDAR point, a laser ray is generated from the LiDAR sensor center to this point. All the grid cells intersected with this ray are visited and this ray will end by the first grid cell containing at least one LiDAR point. This grid cell is then marked as occupied. The other visited empty grid cells are marked as free. Finally, this 3D grid is marked by three states, \emph{unknown}, \emph{free}, and \emph{occupied}, forming a sparse representation of occupancy feature in 3D grid cells.

The encoding method of occupancy feature in MASS is a slightly modified version based on the aforementioned visibility feature. The occupancy feature utilized in MASS is called as \emph{observability} feature encoded in the dense 2D top-view form.
The observability is slightly different compared with the aforementioned visibility.
First, it leverages pillars to take the place of the voxel representation.
Second, the three states in visibility feature are discarded and the accumulated ray passing number is used to encode occupancy.
Finally, we obtain a densely encoded occupancy feature map on the top view.
The key differences between the observability and  visibility features are illustrated in Fig.~\ref{fig:generatation_procedure}. While the observability  depicts the number of the laser rays intersected with its corresponding pillar for each grid cell, the visibility feature encodes each individual voxel, marking it as \emph{unknown}, \emph{free}, or \emph{occupied}. The observability feature is therefore a \textit{dense} encoding of the environment.
\begin{figure}[t]
\begin{center}
\includegraphics[width=0.95\columnwidth]{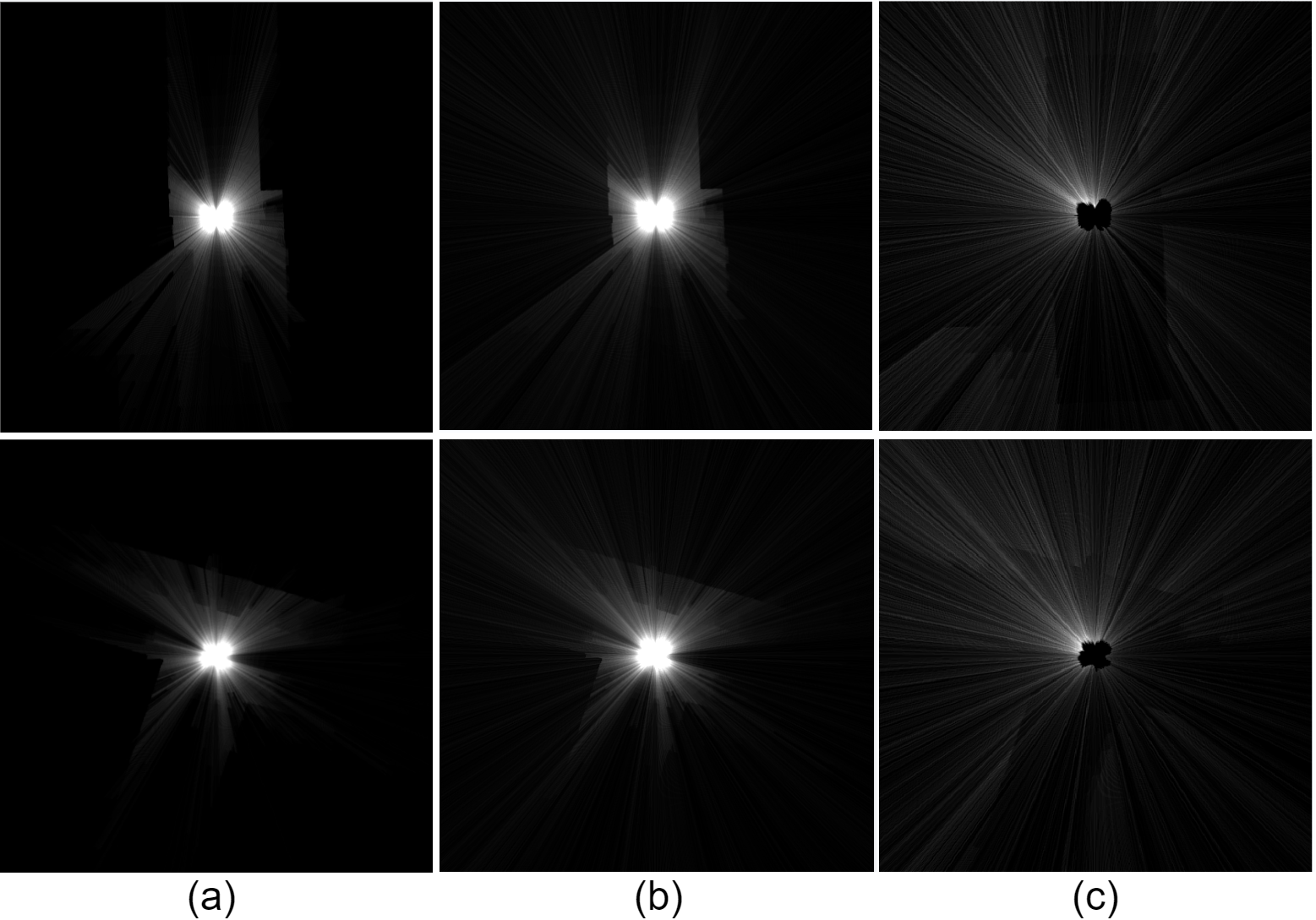}
\end{center}
\vskip-3ex
\caption{Two samples under noise condition SNR (Signal Noise Ratio) = $10$ with random generated noise points, where (a) indicates the observability without generated noise points, (b) indicates the observability with generated noise points, and (c) indicates the absolute difference between (a) and (b).
}
\label{fig:noise}
\vskip-3ex
\end{figure}

We further investigate the tolerance of the observability feature against random noise. We set the Signal Noise Ratio (SNR) condition to SNR $= 10$ and compare the observability feature \textit{without} noise disturbance  in Fig.~\ref{fig:noise}(a), to its counterpart with noise disturbance under the control condition in Fig.~\ref{fig:noise}(b). The impact of noise is further highlighted in Fig.~\ref{fig:noise}(c), which depicts the absolute difference between the corrupted and noise-free variants. The observability has been increased on unknown region where there is no LiDAR point under the attack of the additional noise. Due to the unbalanced ratio between objects such as \emph{building}, which occupies a significant portion of the top-view scene and is not observable, additional random noise will add more points for that part according to the category-wise points ratio and thereby increase the number of laser rays passing through the grid cell that belongs to the road-related region, which makes the difference reasonable.
\begin{figure*}
\begin{center}
\includegraphics[width=2.\columnwidth]{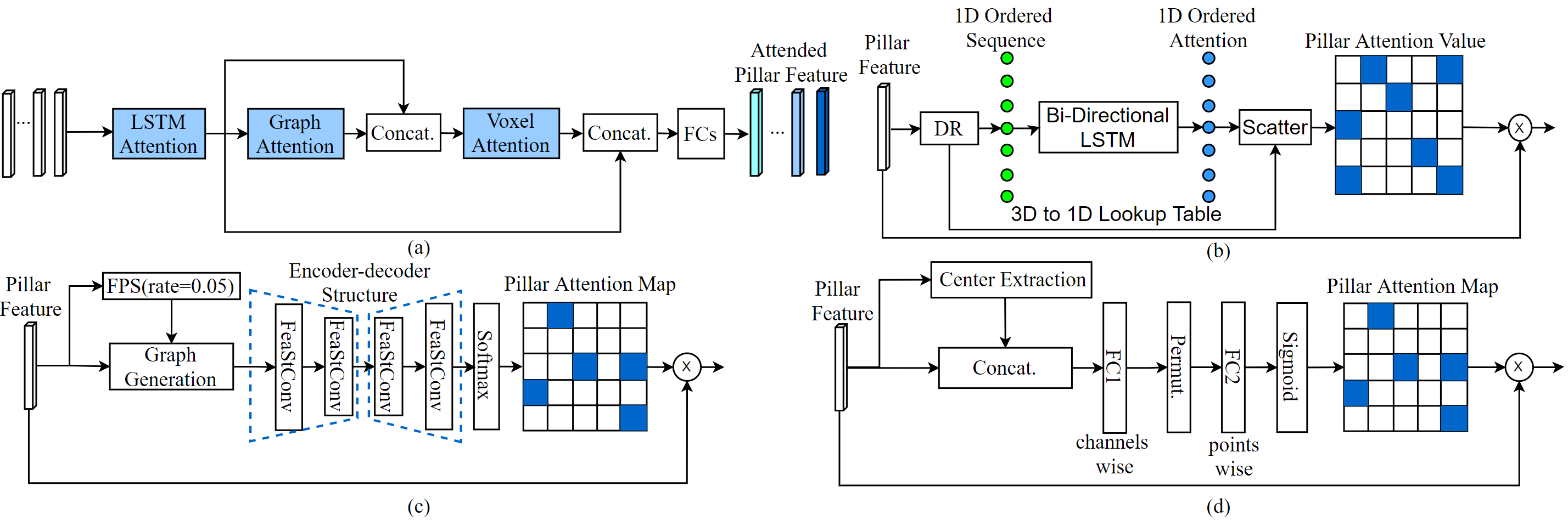}
\end{center}
\vskip-3ex
\caption{Multi-attention (MA) mechanisms proposed in our work, where (a) depicts the general workflow of MA, (b) depicts the dimension reduction (DR) based LSTM attention, (c) depicts the attention generation workflow of key-node based graph attention, and (d) introduces pillar attention according to~\cite{tanet}.}  

\label{fig:ma_mechanism}
\end{figure*}
\subsection{LSTM Attention with Dimension Reduction Index Embedding (DR LSTM)}
PointNet~\cite{pointnet} is mainly built by fully connected layers which cannot preserve locality compared with convolutional layers from 2D CNN, which becomes a challenge for feature extraction of 3D point cloud. 
To alleviate this issue, we leverage an LSTM-based model, since a 3D LiDAR point cloud can be viewed as a sequence and LSTM aggregates the locality features according to the distance.
We therefore propose to leverage LSTM attention with spatial embedding on 3D point cloud data.
We use a bidirectional LSTM to harvest locality-preserving features in a high-dimensional feature space according to distance encoded by spatial embedding to generate a local-preserved attention map, which we now explain. 
In order to implement the sequence processing method, position embedding is required for the pillar-level node to generate the input for the bidirectional LSTM. 
First, we reduce the dimensionality of our data by using principle component analysis (PCA) for dense top-view semantic segmentation and local preserve projection (LPP) for 3D object detection due to different memory consumption of different tasks, leading to a 1D spatial embedding.
In this way, we are able to generate 1D ordered sequence for the input of the bidirectional LSTM attention. 
After obtaining this position embedding, pillar-level nodes are sorted according to the position embedding to form an ordered sequence. The resulting sequence represents the whole input pillar set in the high-level feature space. This ordered sequence is then fed into the bidirectional LSTM module to generate the attention map. 
\subsection{Key-node based Graph Attention}
Since 3D point cloud is relatively noisy~\cite{han2017review}, only few points contain significant clues for dense top-view semantic segmentation. Thereby, we propose a novel key-node based graph attention mechanism which propagates relevant cues from key-nodes to the other nodes. The representative node for each pillar is generated through a \emph{max} operation among all points inside a non-empty pillar. Farthest Point Selection (FPS) is leveraged to generate the key-node set in a high-level representation whose information is used to enrich the information of other pillar-level nodes utilizing graph convolution according to the distance in the high-level representation space. A fully connected graph between the key-node set and the original input set is built for the graph attention generation. 

\textbf{Feature-Steered Graph Convolution.}
To generate better attention maps, we further leverage feature-steered graph convolution (FeaStConv)~\cite{feastnet} to form a graph attention model in an encoder-decoder structure. Our motivation behind this step is the translation invariance facilitated by FeaStConv, which works particularly well in 3D shape encoding.
Graph convolution enables long-chain communication and information flow between the nodes.
We now describe the basic workflow of FeaStConv adopted to our dense semantic segmentation task.

First, neighbourhood information is encoded in a fully connected graph composed of nodes and edges, which are pillar-level nodes and the neighbourhood distance, while the neighbourhood weights of each node are learned in an end-to-end fashion. 
This procedure is designed to simulate the workflow of convolutional layer, which has the capability to aggregate features inside a specific field of view defined by a neighbourhood distance. Second, an additional soft alignment vector proposed in FeaStConv~\cite{feastnet} is leveraged in order to introduce robustness against variations in degree of nodes. The soft alignment parameters are also learned end-to-end. Finally, the desired feature is aggregated through a \emph{sum} operation over the soft aligned, weighted neighbourhood nodes inside the defined neighbourhood.

In FeaStConv, soft-alignment vector $p_m(x_i, ~x_j)$ for node $i$ scales $m$-th weight matrix $W_m$ for feature aggregation as depicted in the following:
\begin{equation}
\begin{split}
\centering
    y_i = b+ \sum_{m=1}^M\frac{1}{|N_i|}\sum_{j\in N_i}p_m(x_i,x_j)W_m x_j,
\end{split}
\end{equation}
\begin{equation}
\begin{aligned}
\centering
    p_m(x_i, x_j) \propto \mathrm{exp}(u_m^T(x_j-x_i)+c_m),
\end{aligned}
\end{equation}
where $u_m, ~v_m$, and $c_m$ are parameters of linear transformation that can be directly learned during the training process with the condition $\sum_{m=1}^M p_m(x_i, x_j)=1$. $x_i$ indicates the node feature of point $i$. $N_i$ indicates the neighbourhood of point $i$ leveraged to aggregate features.

\textbf{Attention Generation Model Structure.}
Owing to the sparsity of 3D point cloud, only a small portion of the points is vital to our task. In the proposed graph attention generation mechanism, the key nodes are selected by utilizing FPS. A bidirectional graph is constructed between the key-node set and the original input set in a fully connected style.
In contrast to graph generated through the K-nearest neighbour method that only considers several nearby nodes, the fully connected graph constructed in our work is able to link key nodes to all other nodes and thereby captures long-range multi-step dependencies.
An encoder-decoder structure constructed based on FeaStConv is utilized to generate graph attention. 
This attention generation procedure is illustrated in Fig.~\ref{fig:ma_mechanism}(a).

\subsection{Pillar Attention}
Pillar attention aggregates features among points inside a pillar and also among channels aiming at the high-level representation to form the attention maps, as done in~\cite{tanet} for 3D object detection.
Our MA leverages this attention to aggregate cues among points and channels to improve the performance of dense top-view semantic segmentation. 
The procedure of generating such attention maps is now detailed.

After the extraction of the pillar center coordinates, the original pillar feature is concatenated with these extracted center coordinates. Then a channel-wise fully connected layer with \emph{ReLU} activation is utilized, which has a decreasing channel number in order to aggregate features along the channel axis.

Then, output features from the first fully connected layer are permuted and fed into another fully connected layer to aggregate features among all the points inside a pillar. The desired pillar attention map is generated based on the output of the second fully connected layer utilizing the \emph{Sigmoid} function.
Channel-wise feature aggregation and point-wise feature aggregation are realized through this procedure. Assuming $N$ is the total number of points inside a pillar, $C$ is the input channel number, and $P$ is the total number of pillars, the first fully connected layer reduces the channel number of pillar features to $1$ and changes the size of the feature map as $(P,~N,~1)$, whereas the second fully connected layer reduces the point number inside a pillar to $1$ and changes the size to $(P,~1,~1)$. Finally, this attention map can be multiplied with the input pillar-level feature as depicted in Fig.~\ref{fig:ma_mechanism}(c).
 
\subsection{Multi-Attention Model}
Our complete frameworks overs three types of attention mechanisms described previously.
In this section, we describe the interplay of the three techniques, with the complete fusion model structure provided in Fig.~\ref{fig:ma_mechanism}(d).
As it comes to the attention order, we first execute the LSTM attention, followed by the graph attention, and, finally, the pillar attention.
The weighted pillar level feature after the LSTM attention is concatenated with the input of the pillar attention module and then passed through several fully connected layers.

\textbf{A note on attention order.} The order of these three attention blocks is determined by the range of the feature aggregation. As aforementioned, the LSTM attention is able to conserve \textit{locality} since the pillars with different distances contribute differently. Graph attention is a \textit{global} attention which propagates important cues from key node to the other nodes. Pillar attention is also a \textit{local} attention generating self-attention, which is more local than LSTM attention. MASS follows a \textit{local-global-local} order to encourage incremental feature enhancement among different attentions. For example, if the global attention is not in the middle, then the first two local attentions will be redundant. The pillar attention is placed at the end following~\cite{tanet}.
The illustration of ablation experimental results in Sec.~\ref{sec:analysis_of_mass} also verifies the analysis.
\subsection{Loss Function.}
We use weighted cross entropy loss to optimize our model on the dense top-view semantic segmentation task. 
The weights for different classes are set according to their statistical distribution. The loss function is therefore formulated as:
\begin{equation}
    \centering
        \mathcal{L}_{\mathrm{seg}} = -\frac{1}{M}\sum_{i=1}^{M}(\lambda y_i\mathrm{log}\hat{y}_i+(1-\lambda)(1-y_i)\mathrm{log}(1-\hat{y}_i)),
\end{equation}
where $y_i$ and $\hat{y}_i$ indicates the ground truth and \emph{Softmax} probability estimated for $i$-th grid cell on the top view, For sparse supervision, $y_i$ indicates the sparse top-view label, while for dense supervision, $y_i$ is the dense top-view label. $\lambda$ is the class-specific weight, and $M$ denotes the number of labeled grid cell on the top view.
The weight coefficient is chosen as $2$ for \emph{vehicle}, and $8$ for \emph{pedestrian}, \emph{two-wheel}, and \emph{rider} in the \emph{Dense Train} mode. For the \emph{Sparse Train} mode, the weight coefficient of \emph{vehicle} is changed to $5$. For other classes, the weight coefficient is set as $1$ to calibrate a good balance among different classes. We remove the channel to predict unlabeled location to force the model make a decision among all the known classes for the unlabeled part marked as white region indicated by the first column of Fig.~\ref{fig:nusc_visual}. The white region on the output is not the unlabeled category. It indicates the unobserved region after filtering by the observation mask as shown in the last column of Fig.~\ref{fig:nusc_visual}. In this way, a dense top-view semantic segmentation result can be achieved. Note that the final prediction result of our proposed approach is a dense semantic segmentation map on the top view.

For the cross-task efficacy verification of our model on 3D object detection, we introduce the loss function as the depicted in the following.
According to the output of SSD~\cite{liu2016ssd}, the loss to train 3D object detection model is composed of localization regression loss and object classification loss. Bounding box localization loss is defined in the following:
\begin{equation}
\centering
    L_{\mathrm{loc}}=\sum_{b\in(x,y,z,w,l,h,\theta)}\mathrm{SmoothL1}(\Delta b),
\end{equation}
with
\begin{equation}
\begin{split}
\centering
    \Delta x&=\frac{x^{g_t} - x^a}{d^a}, \Delta y=\frac{y^{g_t}-y^a}{d^a}, \Delta z=\frac{z^{g_t}-z^a}{h^a},\\
    \Delta w&=\mathrm{log}\frac{w^{g_t}}{w^a}, \Delta l=\mathrm{log}\frac{l^{g_t}}{l^a}, \Delta w=\mathrm{log}\frac{h^{g_t}}{h^a},\\
    \Delta \theta &= \mathrm{sin}(\theta^{g_t}-\theta^a),
\end{split}
\end{equation}
where $x$, $y$, and $z$ denotes three coordinates of bounding box center in 3D space. $w$, $h$, and $l$ denote width, height, and length of the 3D bounding box. $\theta$ indicates the orientation angle of the 3D bounding box. $x^{g_t}$ and $x^a$ denote the ground truth of coordinate $x$ and predicted coordinate $x$ with $d^a=\sqrt{(w^a)^2+(l^a)^2}$.
Cross entropy loss is leveraged to regress bounding box angle on several discretized directions represented by $L_\mathrm{dir}$.
Focal loss is used for the object classification loss as depicted in the following:
\begin{equation}
\centering
    L_{\mathrm{cls}}=-\alpha^a(1-p^a)^\gamma \mathrm{\mathrm{log}}(p^a),
\end{equation}
where $p^a$ is the anchor class probability and the setting of $\alpha$ and $\gamma$ are chosen as $0.25$ and $2$ separately, which are the same as the setting in PointPillars~\cite{pointpillars}.
The total loss is depicted in the following, where $N_{\mathrm{pos}}$ is the total number of the positive anchors and the weights for each loss $\beta_{\mathrm{loc}}$, $\beta_{\mathrm{cls}}$, and $\beta_{\mathrm{dir}}$ are chosen as $2$, $1$, and $0.2$, individually.
\begin{equation}
\centering
    L=\frac{1}{N_{\mathrm{pos}}}(\beta_{\mathrm{loc}}L_{\mathrm{loc}}+\beta_{\mathrm{cls}}L_{\mathrm{cls}}+\beta_{\mathrm{dir}}L_{\mathrm{dir}}).
\end{equation}

\section{Experimental Setups and Datasets}

Using prominent datasets, we validate our approach for (1) our primary task of dense top-view semantic segmentation and (2) 3D object detection, in order to test the generalization of our approach to other 3D vision tasks.
The datasets utilized in our experiments, the label generation approach, evaluation metrics, and setups are now presented in detail. For semantic segmentation, MASS is compared with the method also focusing on dense top-view understanding, since other methods such as GndNet~\cite{gndnet} aiming at predicting semantic segmentation label for each sparse LiDAR point, have a different ground truth modality compared with our work. 

\begin{table*}[!t]
\centering
\caption{Quantitative results on the SemanticKITTI dataset~\cite{semantic_kitti}, where \emph{Occ} indicates occupancy feature, \emph{P} indicates pillar attention, \emph{L} indicates DR LSTM attention, and \emph{G} indicates graph attention. }
\label{tab:experiments_overall}
\scalebox{0.96}{\begin{tabular}{ll | c | cccccccccccc}
\toprule
Mode
& Method
& \rotatebox{90}{\textbf{mIoU} [\%]}
& \rotatebox{90}{\semcolor[othervehicle] vehicle}
& \rotatebox{90}{\semcolor[person] person}
& \rotatebox{90}{\semcolor[motorcycle] two-wheel}
& \rotatebox{90}{\semcolor[motorcyclist] rider}
& \rotatebox{90}{\semcolor[road] road}
& \rotatebox{90}{\semcolor[sidewalk] sidewalk}
& \rotatebox{90}{\semcolor[parking] other-ground}
& \rotatebox{90}{\semcolor[building] building}
& \rotatebox{90}{\semcolor[fence] object}
& \rotatebox{90}{\semcolor[vegetation] vegetation}
& \rotatebox{90}{\semcolor[trunk] trunk}
& \rotatebox{90}{\semcolor[terrain] terrain} \\
\midrule
\multirow{4}{*}{\shortstack[l]{Sparse Train\\Sparse Eval}}
& Bieder~\textit{et al.}~\cite{bieder2020exploiting}           & \textbf{39.8} & 69.7 & 0.0  & 0.0  & 0.0  & 85.8 & 60.3 & 25.9 & 72.8 & 15.1 & 68.9 & 9.9  & 69.3 \\
& Pillar~\cite{fei2021pillarsegnet}                 & \textbf{55.1} & 79.5 & 15.8 & 25.8 & 51.8 & 89.5 & 70.0 & 38.9 & 80.6 & 25.5 & 72.8 & 38.1 & 72.7  \\
& Pillar + Occ~\cite{fei2021pillarsegnet}  & \textbf{55.3} & 82.7 & 20.3 & 24.5 & 51.3 & 90.0 & 71.2 & 36.5 & 81.3 & 28.3 & 70.4 & 38.5 & 69.0 \\
& Pillar + Occ + P & \textbf{57.5} &  85.1& 24.7 & 16.9 & 60.1 & 90.7 & 72.9 & 38.3 & 82.9 & 30.1 & 80.4 & 35.4 & 72.8   \\
& Pillar + Occ + LP & \textbf{57.8} &  85.9& 24.2 & 18.3 & 57.6 & 91.3 & 74.2 & 39.2 & 82.4 & 29.0 & 80.6 & 38.0 & 72.9   \\
& Pillar + Occ + LGP  & \textbf{58.8} &  85.8& 34.2 & 26.8 & 58.5 & 91.3 & 74.0 & 38.1 & 82.2 & 28.7 & 79.5 & 35.7 & 71.3 \\

\midrule
\multirow{4}{*}{\shortstack[l]{Sparse Train\\Dense Eval}}
& Bieder~\textit{et al.}~\cite{bieder2020exploiting}          & \textbf{32.8} & 43.3 & 0.0 & 0.0 & 0.0 & 84.3 & 51.4 & 22.9 & 54.7 & 10.8 & 51.0 & 6.3 & 68.6  \\
& Pillar~\cite{fei2021pillarsegnet}                 & \textbf{37.5} & 45.1 & 0.0 & 0.1 & 3.3 & 82.7 & 57.5 & 29.7 & 64.6 & 14.0 & 58.5 & 25.5 & 68.9   \\
& Pillar + Occ~\cite{fei2021pillarsegnet}  & \textbf{38.4} & 52.5 & 0.0 & 0.2 & 3.0 & 85.6 & 60.1 & 29.8 & 65.7 & 16.1 & 56.7 & 26.2 & 64.5   \\
& Pillar + Occ + P & \textbf{40.9} &  53.3& 11.3 & 13.1 & 7.0 & 83.6 & 60.3 & 30.2 & 63.4 & 15.7 & 61.4 & 24.6 & 67.2   \\
& Pillar + Occ + LP  & \textbf{41.5} &  57.3& 11.3 & 9.5 & 10.4 & 85.5 & 60.1 & 31.2 & 64.6 & 16.9 & 59.5 & 25.3 & 66.8   \\
& Pillar + Occ + LGP  & \textbf{40.4} &  55.8& 10.8 & 14.1 & 9.3 & 84.5 & 58.6 & 26.8 & 62.4 & 15.2 & 59.2 & 26.3 & 62.3   \\
\midrule
\multirow{3}{*}{\shortstack[l]{Dense Train\\Dense Eval}}
& Pillar~\cite{fei2021pillarsegnet}                 & \textbf{42.8} & 70.3 & 5.4 & 6.0 & 8.0  & 89.8 & 65.7 & 34.0 & 65.9 & 16.3 & 61.2 & 23.5 & 67.9   \\
& Pillar + Occ~\cite{fei2021pillarsegnet}  & \textbf{44.1} & 72.8 & 7.4 & 4.7 & 10.2 & 90.1 & 66.2 & 32.4 & 67.8 & 17.4 & 63.1 & 27.6 & 69.2   \\
& Pillar + Occ + P  & \textbf{44.9} & 72.1 & 6.8 & 6.2 & 9.9 & 90.1 & 65.8 & 37.8 & 67.1 & 18.8 & 68.1 & 24.7 & 71.4   \\
& Pillar + Occ + LP  & \textbf{44.8} &  73.0& 7.8 & 6.1 & 10.6 & 90.6 & 66.5 & 33.7 & 67.6 & 17.7 & 67.6 & 25.5 & 70.4   \\
& Pillar + Occ + LGP  & \textbf{44.5} &  73.2 & 6.5 & 6.5 & 9.5 & 90.8 & 66.5 & 34.9 & 68.0 & 18.8 & 67.0& 22.8 & 70.0   \\
\bottomrule
\end{tabular}}
\end{table*}

\subsection{Datasets}
\textbf{SemanticKITTI.}
Our MASS model is first trained and evaluated on the SemanticKITTI dataset~\cite{semantic_kitti} providing semantic annotations for a subset of the KITTI odometry dataset~\cite{kitti} together with pose annotations.
We follow the setting of~\cite{semantic_kitti}, using sequences 00-07 and sequences 09-10  as the training set containing $19130$ LiDAR scans, while the sequence 08 is used as the evaluation set containing $4071$ LiDAR scans.
As in~\cite{bieder2020exploiting}, our class setup merges $19$ classes into $12$ classes (see Table~\ref{tab:experiments_overall}) to facilitate fair comparisons. The class mapping is defined in the following.
\emph{Car}, \emph{truck}, and \emph{other-vehicle} are mapped to \emph{vehicle}, meanwhile the classes \emph{motorcyclist} and \emph{bicyclist} are mapped to \emph{rider}.
The classes \emph{bicycle} and \emph{motorcycle} are mapped to \emph{two-wheel}, whereas the classes \emph{traffic-sign}, \emph{pole}, and \emph{fence} are mapped to \emph{object}.
The classes \emph{other-ground} and \emph{parking} are mapped to \emph{other-ground}, while \emph{unlabeled} pixels are not considered during the loss calculation which means the supervision is only executed on labeled grid cells to achieve dense top-view semantic segmentation prediction.

\textbf{nuScenes-LidarSeg.} The novel nuScenes-LidarSeg dataset~\cite{nuscenes} covers semantic annotation for each LiDAR point for each key frame with $32$ possible classes.
Overall, $1.4$ billion points with annotations across $1000$ scenes and $40,000$ point clouds are contained in this dataset.
The detailed class mapping is defined as follows. \emph{Adult}, \emph{child}, \emph{construction worker}, and \emph{police officer} are mapped as \emph{pedestrian}.
\emph{Bendy bus} and \emph{rigid bus} are mapped as \emph{bus}. The class mapping for \emph{barrier}, \emph{car}, \emph{construction vehicle}, \emph{motorcycle}, \emph{traffic cone}, \emph{trailer}, \emph{truck}, \emph{drivable surface}, \emph{other flat}, \emph{sidewalk}, \emph{terrain}, \emph{manmade}, and \emph{vegetation} are identical.
The other classes are all mapped to \emph{unlabeled}.
Thereby, we study with $12$ classes (see Table~\ref{tab:experiments_nusc}) for dense semantic understanding on nuScenes-LidarSeg.
The supervision mode is the same as that on SemanticKITTI as aforementioned. 

\begin{table*}[!t]
\centering
\caption{Quantitative results on the nuScenes dataset~\cite{nuscenes}. The order of the three kinds of attention is indicated in the brackets. For example, LGP indicates the order of first DR LSTM attention, second graph attention, and finally pillar attention.}
\scalebox{0.95}{\resizebox{\textwidth}{26mm}{
\label{tab:experiments_nusc}
\begin{tabular}{l | c | c | cccccccccccccccc}
\toprule
Mode
& Method
& \rotatebox{90}{\textbf{mIoU} [\%]}
& \rotatebox{90}{\semcolor[barrier_n] barrier}
& \rotatebox{90}{\semcolor[bicycle_n] bicycle}
& \rotatebox{90}{\semcolor[bus_n] bus}
& \rotatebox{90}{\semcolor[car_n] car}
& \rotatebox{90}{\semcolor[c_v_n] const-vehicle}
& \rotatebox{90}{\semcolor[motorcycle_n] motorcycle}
& \rotatebox{90}{\semcolor[pedestrian_n] pedestrian}
& \rotatebox{90}{\semcolor[cone_n] cone}
& \rotatebox{90}{\semcolor[trailer_n] trailer}
& \rotatebox{90}{\semcolor[truck_n] truck}
& \rotatebox{90}{\semcolor[drive_n] drivable}
& \rotatebox{90}{\semcolor[other_flat_n] other-flat} 
& \rotatebox{90}{\semcolor[sidewalk_n] sidewalk} 
& \rotatebox{90}{\semcolor[terrain_n] terrain} 
& \rotatebox{90}{\semcolor[manmade_n] manmade} 
& \rotatebox{90}{\semcolor[vegetation_n] vegetation} \\
\midrule
\multirow{2}{*}{\shortstack[l]{Dense Train\\Dense Eval}}
& Pillar & \textbf{22.7} & 10.8 & 0.0  & 5.3  & 1.6  & 6.0 & 0.0 & 0.0 & 0.8 & 19.59 & 0.8 & 83.4  & 35.5 & 45.0 & 52.3 & 48.5 & 54.3 \\
& MASS & \textbf{30.4} & 25.3 & 0.0 & 20.7 & 25.2 & 14.4 & 0.0 & 3.3 & 1.4 & 26.8 & 14.9 & 86.8 & 46.0 & 50.4 & 55.7 & 55.9 & 61.0\\
\midrule
\multirow{2}{*}{\shortstack[l]{Noise Ablation}}&Pillar & \textbf{15.9} & 2.6 & 0.0 & 0.9 & 0.2  & 0.8 & 0.0 & 0.0 & 0.0 & 6.1 & 0.0 & 72.5 & 10.1 & 29.9 & 40.7 &  45.0 & 45.1\\
& MASS & \textbf{29.8} & 22.1 & 0.0 & 23.0 & 26.7 & 15.9 & 0.5 & 2.5 & 1.2 & 24.9 & 16.4 & 84.5 & 42.9 & 47.9 & 53.1 & 56.3 & 58.2\\
\midrule
\multirow{3}{*}{\shortstack[l]{Order Ablation}}
&MASS (GLP) & \textbf{26.3} &22.6& 0.0 & 19.6 & 22.3 & 11.6 & 0.0 & 1.4 & 0.2 & 23.9& 9.1 & 83.4 & 34.9 & 42.4 & 40.0& 51.3 & 57.3 \\
&MASS (LPG) & \textbf{28.9} & 22.9 & 0.0 & 21.5 & 23.4 & 11.4 & 0.1 & 2.0 & 0.8 & 22.6& 11.7 & 85.9 & 43.3& 48.6& 53.9& 54.7 & 59.9\\
&MASS (PLG) & \textbf{30.2} &24.5& 0.0 & 20.7 & 28.0 & 13.1& 0.0& 3.5 & 2.1 & 25.1 & 15.4 & 86.3 & 45.8 & 49.3 & 54.5 & 55.2 & 60.4\\
 & MASS (LGP) & \textbf{30.4} & 25.3 & 0.0 & 20.7 & 25.2 & 14.4 & 0.0 & 3.3 & 1.4 & 26.8 & 14.9 & 86.8 & 46.0 & 50.4 & 55.7 & 55.9 & 61.0\\
\bottomrule
\end{tabular}}}
\vskip-3ex
\end{table*}

\textbf{KITTI 3D object detection dataset.}
To verify the cross-task generalization of our MA model, we use the KITTI 3D object detection dataset~\cite{kitti}.
It includes $7481$ training frames and $7518$ test frames with $80256$ annotated objects.
Data for this benchmark contains color images from left and right cameras, 3D point clouds  generated through a Velodyne LiDAR sensor, calibration information, and training annotations.

\subsection{Sparse Label Generation}
The point cloud is first rasterized into grid cells representation on the top view in order to obtain cell-wise semantic segmentation annotations through a weighted statistic analysis for the occurrence frequency of each class inside each grid cell. The number of points inside each grid cell for each class is counted at first.
The semantic annotation $k_i$ for grid cell $i$ is then calculated through a weighted $\mathrm{argmax}$ operation depicted in the following:
\begin{equation}
\label{eq:argmax}
\centering
k_{i} = \underset{k\in[1,K]}{\mathrm{argmax}} \left( w_k  n_{i,k}\right),
\end{equation}
where $K$ is the total class number, $n_{i,k}$ denotes the number of points for class $k$ in grid cell $i$, and $w_k$ is the weight for class $k$.

For traffic participant classes including \emph{vehicle}, \emph{person}, \emph{rider}, and \emph{two-wheel}, the weight is chosen as $5$ according to the class distribution mentioned in~\cite{bieder2020exploiting}.
Since the aforementioned \emph{unlabeled} class is discarded during training and evaluation, in order to achieve fully dense top-view semantic segmentation, the weight for this label is then set to $0$.
The weight for the other classes is set as $1$ to alleviate the heavy class-distribution imbalance according to the statistic distribution of point numbers of different classes detailed in~\cite{bieder2020exploiting}.
Grid cells without any assigned points are finally annotated as \emph{unlabeled} and loss is not calculated on them.

\subsection{Dense Label Generation}
Dense top-view semantic segmentation ground truth is generated to achieve a more accurate evaluation and can be also utilized to train the MASS network to facilitate comparability.
The multi-frame point cloud concatenation procedure leveraged for label generation only considers LiDAR point clouds belonging to the same scene. The generation procedure of dense top-view semantic segmentation ground truth is described in detail in the following.

First, a threshold of ego-pose difference is defined as twice of the farthest LiDAR point distance $d$ to select nearby frames for each frame in the dataset.
When the ego pose distance between the current frame and a nearby frame, $|\Delta p_x|$, is smaller than the threshold $d$, this nearby frame is selected into the candidate set to densify the semantic segmentation ground truth. The densification process is achieved through unification of coordinates based on the pose annotation for each nearby frame. Only static objects of the nearby frames are considered, since dynamic objects can cause aliasing in this process.

\subsection{Evaluation Metrics}

The evaluation metrics for dense top-view semantic segmentation is Intersection over Union (IoU) and mean of Intersection over Union (mIoU) defined in the following equation:
\begin{equation}
\centering
    \mathrm{IoU}_i = \frac{A_i\cap B_i}{A_i \cup B_i}, \mathrm{mIoU} = \frac{1}{K}\sum_{i=1}^K \mathrm{IoU}_i,
\end{equation}
where $A_i$ denotes pixel number with the ground truth for class $i$, $B_i$ denotes the pixel number with predicted semantic segmentation labels for class $i$, and $K$ indicates the total class number. For dense top-view semantic segmentation, only visible region is selected for the evaluation procedure.

The evaluation metrics for 3D object detection are Average Precision (AP) and mean Average Precision (mAP) which are defined by the following:
\begin{equation}
\centering
    AP = \sum_{k=1}^n P(k)\Delta r(k),
\end{equation}
where $P(k)$ indicates the precision of current prediction and $\Delta r(k)$ indicates the change of recall.

\subsection{Implementation Details}
In the following, the model setup of the pillar feature net, 2D backbone, data augmentation, and the training loss are described in detail.

\textbf{Pillar Extraction Network Setup.}
First, we introduce the model setup for our primary task of dense top-view semantic segmentation. The given 3D point cloud is cropped on the $x,~y,~z$ axes using the ranges $[-50.0,~50.0]m$, $[-25.0,~25.0]m$, and $[-2.5,~1.5]m$ accordingly, and the pillar size along $x,~y,~z$ directions is defined as $[0.1,~0.1,~4.0]m$.
We set a maximum point number inside each pillar to $20$ in order to receive a fair comparison with the dense top-view semantic segmentation results from~\cite{bieder2020exploiting} on SemanicKITTI~\cite{semantic_kitti}.

For the experiments on nuScenes-LidarSeg \cite{nuscenes}, the range for $x,~y,~z$ is set to $[-51.2, ~51.2]m$, $[-51.2,~51.2]m$, and $[-5,~3]m$, while the pillar size is $[0.2,~0.2,~8.0]m$. The input feature comprises $10$ channels, while the output of the pillar feature net is $64$ channels for both datasets, which is lifted through PonitNet~\cite{pointnet}. Our model is based on OpenPCDet.\footnote{\url{https://github.com/open-mmlab/OpenPCDet}}

Second, we showcase the model setup for verification of the cross-task generalization.
The backbone codebase we use is \textit{second.pytorch}.\footnote{\url{https://github.com/traveller59/second.pytorch.git.}}
The resolution for the $xy$ plane is set as $0.16 m$, the maximum number of pillars is $12000$, and the maximum number of points inside each pillar is $100$. The point cloud ranges of $x,~y,~z$ axes for \emph{pedestrian} are cropped in range $[0,~47.36]m$, $[-19.48,~19.84]m$, $[-2,5,~0.5]m$, whereas the ones for \emph{car} are set as $[0,~69.12]m$, $[-39.68,~39.68]m$, and $[-3,~1]m$.
The resolution on $z$ axis is $3 m$ for \emph{pedestrian} and is $4 m$ for \emph{car}. The input channel number of pillar feature net is $9$ and the output channel number is set as $64$.

\textbf{MA Setup.} 
For graph attention, FPS rate is selected as $0.05$. The encoder-decoder model to generate attention map is composed of $2$ FeaStConv layers in the encoder part and $2$ FeaStConv layers in the decoder part. For LSTM attention, Principle Component Analysis (PCA) is selected for dimension reduction towards dense top-view semantic segmentation and Local Preserving Projection (LPP) is selected for the cross-task efficacy verification of MA due to different memory consumption requirements for different tasks.

\textbf{2D Backbone.} The first 2D backbone introduced here is a Modified UNet (M-UNet) for dense top-view semantic segmentation on SemanticKITTI~\cite{semantic_kitti} and nuScenes-LidarSeg~\cite{nuscenes} datasets. Since our model leverages MA and PonitNet~\cite{pointnet} to encode pillar features and lifts features in high-level representations, the first convolutional block of UNet is discarded due to redundancy, which maps a $3$-channel input to a $64$-channel output, to form the M-UNet leveraged in our approach. M-UNet thereby helps to maintain an efficient model.

The second 2D backbone is for the cross-task efficacy verification of our MA model on 3D object detection on the KITTI 3D detection dataset. This backbone is different from that for dense top-view semantic segmentation. It is composed of a top-down network producing features in increasingly smaller spatial resolutions and an upsampling network that also concatenates top-down features, which is the same as~\cite{pointpillars}.
First, the pillar scatter from PointPillars~\cite{pointpillars} generates a pseudo image on the top view for 2D Backbone's input from aggregated pillars.
A $64$-channel pseudo image is input into the 2D backbone.
The stride for the top-down 2D backbone network is defined as $[2,~2,~2]$ with filter numbers $[64,~128,~256]$ and the upsample stride is defined as $[1,~2,~4]$ with filter numbers $[128,~128,~128]$.

\textbf{Training Setup.}
Weighted cross entropy is leveraged to solve the heavy class imbalance problem. According to the distribution of points for different classes described by~\cite{bieder2020exploiting}, weights for \emph{rider}, \emph{pedestrian}, and \emph{two-wheel} are set as $8$ for loss calculation.
The weight for \emph{vehicle} is set as $2$. For other classes, the weight is set as $1$.
Adam optimizer~\cite{adam} is leveraged in our proposed approach with batch size $2$ and learning rate $0.001$ for $30$ epochs training. The weight decay is set as $0.01$ together with momentum $0.9$. Step scheduler is used with step list $[5, 10, 15, 20, 25, 30]$ for learning rate decay. The parameter amount of each variant of our approach compared with Bieder~\textit{et al.}~\cite{bieder2020exploiting} is shown in Table \ref{tab:experiments_parameters}. Through comparison, it can be found that MA only slightly increases the parameter number of the whole architecture while significantly improving the top-view semantic segmentation performance. Compared to the work from Bieder~\textit{et al.}~\cite{bieder2020exploiting}, our pillar-based approach has a lighter model structure while showing strong efficacy on the dense top-view semantic segmentation task.
\begin{table}[!t]
\centering
\caption{Model parameters and accuracy under dense train dense eval scenario.}
\label{tab:experiments_parameters}
\begin{tabular}{lccc} 
\toprule
Method & \multicolumn{1}{l}{\#Mparams} & \multicolumn{1}{l}{mIoU}& \multicolumn{1}{l}{Bacbone} \\ 
\midrule
Bieder~\textit{et al.}~\cite{bieder2020exploiting} & 35.480M & 39.8 & Xception 65~\cite{chollet2017xception}\\ 
\midrule
Pillar & 7.414M & 55.1 & PillarSegNet\\
Pillar+Occ & 7.415M & 55.3& PillarSegNet\\
Pillar+Occ+P & 7.416M & 57.5& PillarSegNet\\
Pillar+Occ+LP & 7.417M & 57.8& PillarSegNet\\
Pillar+Occ+LGP & 7.418M & 58.8& PillarSegNet\\
\bottomrule
\end{tabular}
\vskip-3ex
\end{table}

\textbf{Data Augmentation.}
Data augmentation for input feature is defined in the following. Let $(x,~y,~z,~r)$ denotes a single point of the LiDAR point cloud, where $x$, $y$, $z$ indicate the 3D coordinates and $r$ represents the reflectance.
Before being passed to the PointNet, each LiDAR point is augmented with the offsets from the pillar coordinates center $(\Delta x_c, \Delta y_c, \Delta z_c)$ and the offsets $(\Delta x_p, \Delta y_p, \Delta z_p)$ between the point and the pillar center.

Then, data augmentation for our main task, dense top-view semantic segmentation, is detailed in the following. Four data augmentation methods are leveraged in order to introduce more robustness to our model for dense top-view semantic segmentation.
First, random world flip along $x$ and $y$ axis is leveraged.
Then, random world rotation with rotation angle range $[-0.785,~0.785]$ is used to introduce rotation invariance to our model.
Third, random world scaling with range $[0.95,~1.05]$ is used for introducing scale invariance and the last one is random world translation.
The world translation standard error, which is generated through normal distribution, is set as $[5,~5,~0.05]$, and the maximum range is set as three times of standard error in two directions.

Finally, data augmentations for cross-task verification of MA on the KITTI 3D dataset~\cite{kitti} are described.
In the training process, every frame of input is enriched with a random selection of point cloud for corresponding classification classes. The enrichment numbers are different for different classes.
For example for \emph{car}, $15$ targets are selected, whereas for \emph{pedestrian} the enrichment number is $0$.
Bounding box rotation and translation are also utilized. Additionally to these, global augmentation such as random mirroring along $x$ axis, global rotating and scaling are also involved. Localization noise is created through a normal distribution $\mathrm{N}(0,~0.2)$ for $x,~y,~z$ axis. The bounding box rotation for each class is limited inside range $[0,~1.57]$ in meter.

\begin{figure*}[!t]
\begin{center}
\includegraphics[width = 0.95\textwidth]{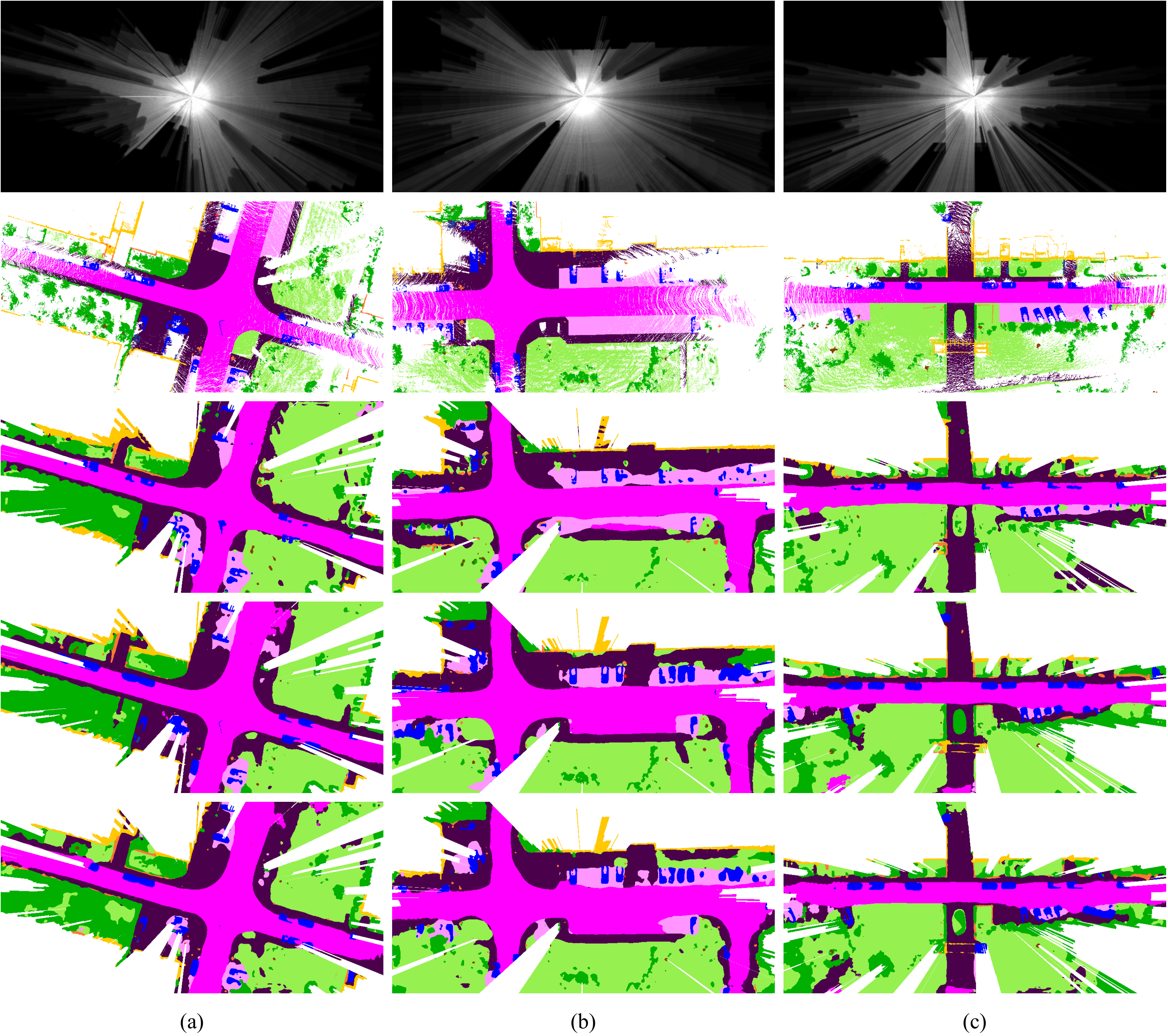}
\end{center}
\vskip-3ex
\caption{Qualitative results on the SemanticKITTI dataset~\cite{semantic_kitti}. 
From top to bottom in each rows, we depict the 2D occupancy map, the ground truth, the prediction from~\cite{bieder2020exploiting}, the prediction from our approach without MA and the prediction of our approach with MA.
The unobservable regions in prediction map were filtered out using the observability map.
In comparison with~\cite{bieder2020exploiting}, our approach without MA and with MA shows more accurate predictions on vehicles and small objects.}
\label{fig:qualitative}
\end{figure*}

\begin{figure*}[!t]
\begin{center}
\includegraphics[width = 0.95\textwidth]{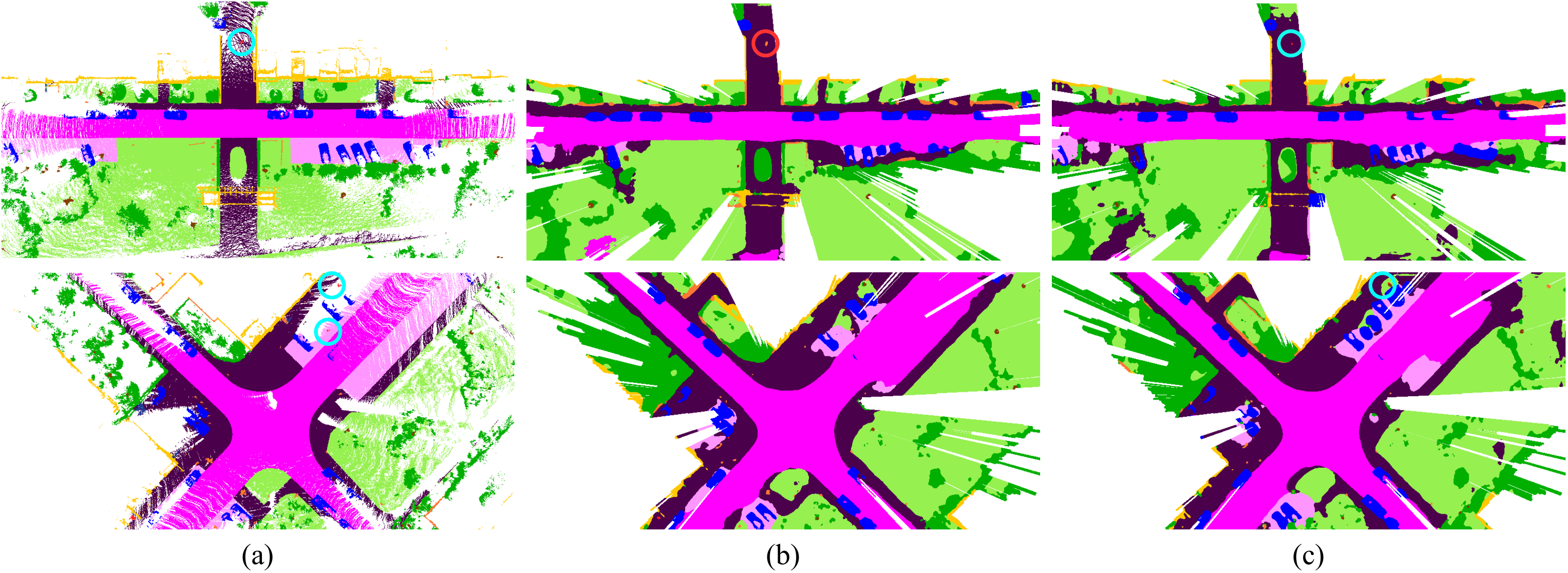}
\end{center}
\vskip-3ex
\caption{A prediction comparison between (b) MASS without MA and (c) MASS with MA, where the ground truth is depicted in (a). Pedestrians in ground truth and true positive predictions are indicated by sky-blue circles, whereas false positive predictions are indicated by red circles.}
\label{fig:comparison}
\end{figure*}

\section{Results and Analysis}
\subsection{Analysis of MASS for Dense Top-View Semantic Segmentation}
\label{sec:analysis_of_mass}
Following the setup of~\cite{bieder2020exploiting}, we consider two training modes and two evaluation modes for dense top-view semantic segmentation: \emph{Sparse Train} and \emph{Dense Train} for training and \emph{Sparse Eval}, and \emph{Dense Eval} for testing. 
\emph{Sparse Train} and \emph{Sparse Eval} take into consideration sparse top-view ground truth obtained through single LiDAR sweep, whereas \emph{Dense Train} and \emph{Dense Eval} utilize the generated dense top-view ground truth to achieve better supervision. 
The evaluation is only considered on visible region on the top-view indicated by the occupancy map and the supervision is only considered on labeled grid cells on the top view to achieve dense predictions. The \emph{Dense Train} experiments are only evaluated by \emph{Dense Eval} approaches, as it has stronger supervision compared with the sparse top-view semantic segmentation ground truth, so that it is not meaningful to evaluate in the \emph{Sparse Eval} mode.

Table~\ref{tab:experiments_overall} summarizes our key findings, indicating, that the proposed pillar-based model surpasses the state-of-the-art grid-map-based method~\cite{bieder2020exploiting} by $15.3\%$ mIoU in the \emph{Sparse Eval} mode and $5.7\%$ mIoU in the \emph{Dense Eval} mode.
Our framework is especially effective for classes with small spatial size such as \emph{person}, \emph{two-wheel}, and \emph{rider}. 
Qualitative results provided in Fig.~\ref{fig:qualitative} also verify the effectiveness of our pillar-based model compared with the previous grid-map-based model. 

We further analyze the significance of the occupancy feature generated through the aforementioned ray casting process and multi-attention (MA) mechanism.
Compared with the model utilizing only pillar features, the added occupancy feature encodes free-space information and brings a performance improvement of $0.9\%$ mIoU in the \emph{Sparse Train Dense Eval} mode and $1.3\%$ in the \emph{Dense Train Dense Eval} mode, indicating that occupancy features can be successfully leveraged for improving dense top-view semantic segmentation. 

Enhancing our framework with the proposed MA mechanism further improves the semantic segmentation results, especially for objects with small spatial size. 
For example, the model with pillar-, DR LSTM- and graph attention gives a $13.9\%$ performance increase for the category \emph{person} in the \emph{Sparse Train Sparse Eval} mode.
Pillar attention firstly brings a $2.2\%$ mIoU boost, the introduction of DR LSTM attention brings a further $0.3\%$ mIoU performance improvement, and finally the graph attention brings a further $1.0\%$ mIoU performance boost compared against the model with occupancy yet without MA.
Overall, our proposed MASS system achieves high performances in all modes. In particular, MASS outperforms the previous state-of-the-art by $19.0\%$ in the \emph{Sparse Train Sparse Eval} mode and $7.6\%$ in the \emph{Sparse Train Dense Eval} mode.

The qualitative results shown in Fig.~\ref{fig:comparison} also verify the capability of MA for detail-preserved fine-grained top-view semantic segmentation.
The model with MA shows strong superiority for the prediction of class \emph{person} indicated by sky-blue circles for ground truth and true positive prediction. The false positive prediction is indicated by red circles. MASS with MA has more true positive predictions and less false positive predictions compared against MASS without MA, demonstrating the effectiveness of MA for dense top-view semantic segmentation.

\begin{figure}[!t]
\begin{center}
\includegraphics[width=1.\columnwidth]{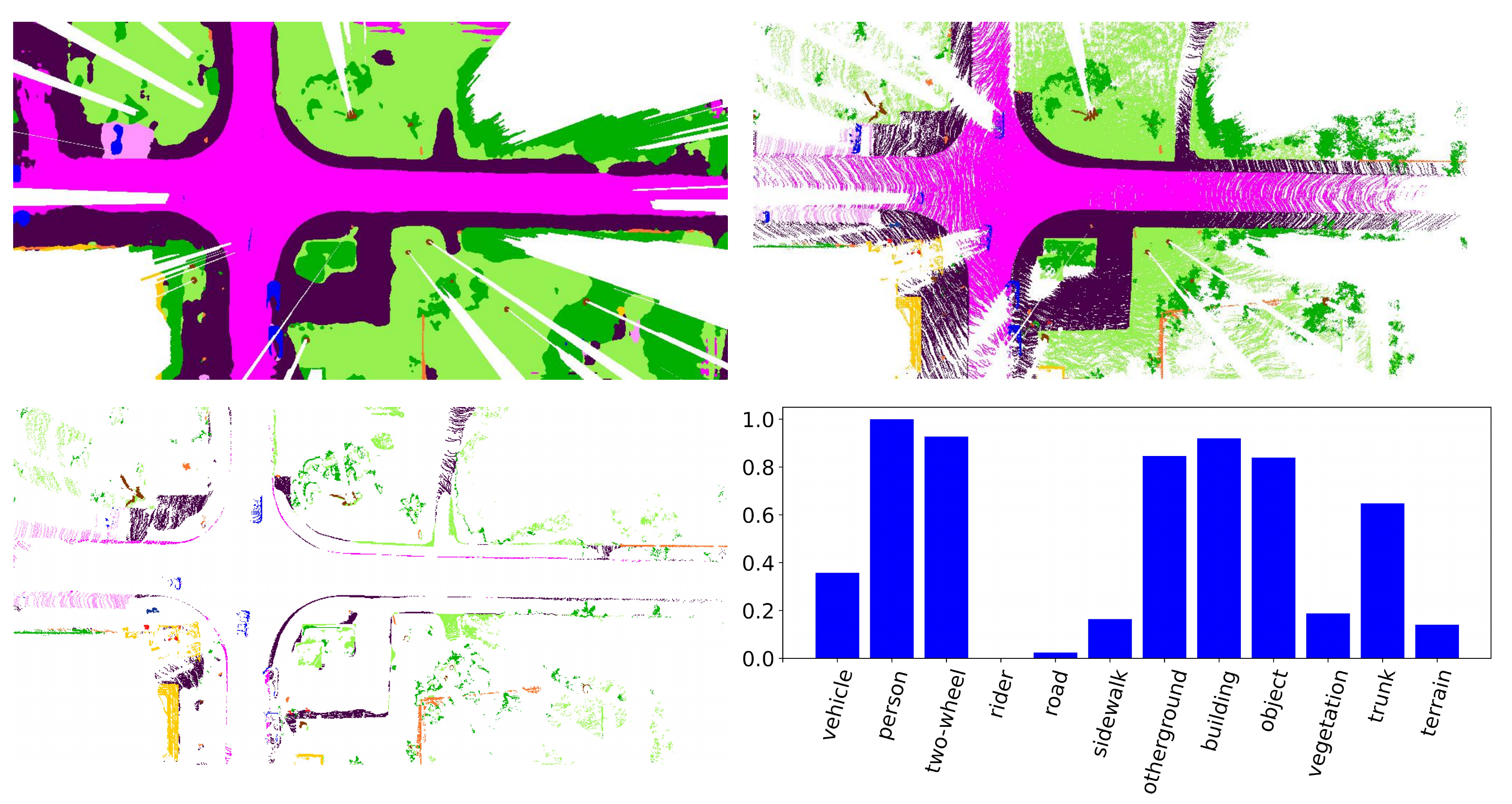}
\end{center}
\vskip-3ex
\caption{Visualization of a failure prediction case. The image on the top left shows the dense top-view semantic segmentation prediction result and the image on the top right shows the dense top-view semantic segmentation ground truth. The image on the bottom left indicates the difference between dense ground truth and prediction, painted according to no-empty grid cell of the ground truth. The sub-figure on the bottom right indicates the analysis of false prediction ratio of this selected frame.}
\vskip-1ex
\label{fig:failure_case}
\end{figure}
A failure case visualization is also provided by our work as depicted in Fig.~\ref{fig:failure_case}. The difference of dense top-view ground truth and prediction result is indicated by the figure on the bottom left for each non-empty grid cell, represented by non-white pixel in the top-view images and painted with color of the correct label for each false prediction on the canvas initialized as white at beginning. Through comparison, \emph{moving car} is found to have a great possibility to be wrongly-predicted due to the unbalanced grid cells number between \emph{moving car} and \emph{stopping car}, since in the dense top-view annotation generation procedure, only static objects are considered to be densified to avoid aliasing. Since in the two datasets leveraged in our work, the movement of each frame is annotated as ego pose change of the data collection car where the LiDAR sensor was mounted on. This issue is possible to be solved if the direction and velocity of moving objects can be obtained relative to the ego pose for a balanced annotation distribution between moving and static objects. Besides, the prediction of edge structure also suffers from low accuracy such as the edge shape object, \emph{building}, in this frame.

\begin{figure*}
\begin{center}
\includegraphics[width = 0.9\textwidth]{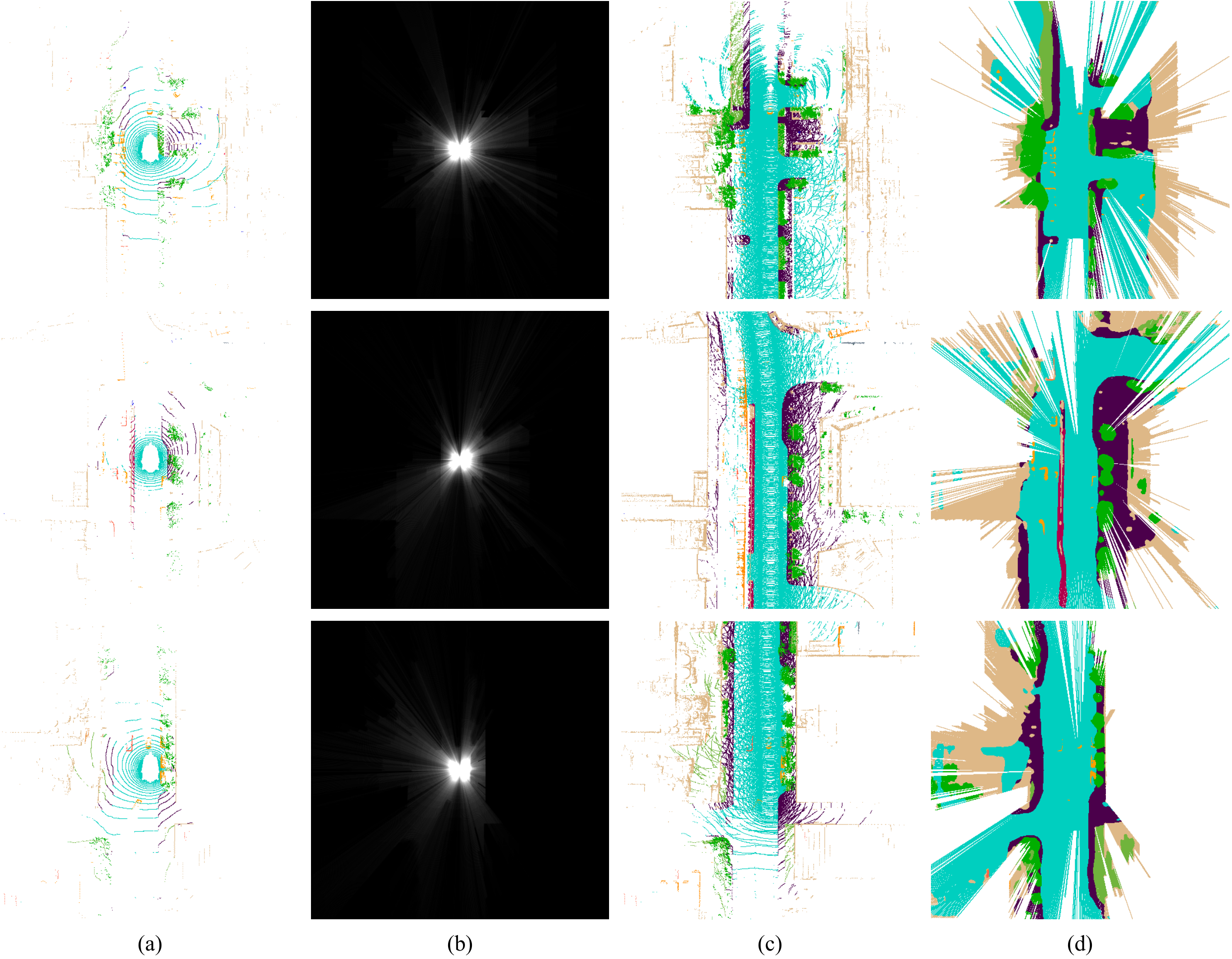}
\end{center}
\vskip-3ex
\caption{Visualization results for dense top-view semantic segmentation prediction on the nuScenes dataset~\cite{nuscenes}. Sparse top-view semantic segmentation ground truth is in column (a), 2D occupancy map is in column (b), dense top-view semantic segmentation ground truth is in column (c) and dense top-view semantic segmentation prediction of MASS is in column (d).}
\label{fig:nusc_visual}
\end{figure*}

In addition to the experiments on SemanticKITTI, we also validate MASS on nuScenes-LidarSeg in order to obtain dense top-view semantic segmentation predictions, which is the first work focusing on this task on nuScenes-LidarSeg based on pure LiDAR data. The visualization results for the dense top-view semantic segmentation prediction, learned on the nuScenes-LidarSeg dataset, are shown in Fig.~\ref{fig:nusc_visual}, where sparse top-view semantic segmentation ground truth, 2D occupancy map, dense top-view semantic segmentation ground truth, and dense top-view semantic segmentation prediction of MASS are illustrated column-wise.
The qualitative results are listed in Table~\ref{tab:experiments_nusc}, where the baseline indicated as \emph{Pillar} achieves $22.7\%$ in mIoU.
Our proposed MASS system with MA and occupancy feature indicated by \emph{MASS} overall significantly boosts the performance, reaching a $7.7\%$ mIoU improvement on nuScenes-LidarSeg, which further verifies the effectiveness of the proposed MA and occupancy feature for dense top-view semantic segmentation. The visualization result of the dense top-view semantic segmentation on the nuScenes-LidarSeg dataset is indicated by Fig.~\ref{fig:nusc_visual}, which shows better understanding of the surrounding environment for the automated vehicle compared with the sparse point-wise semantic segmentation ground truth.

\begin{figure}
\begin{center}
\includegraphics[width=0.9\columnwidth]{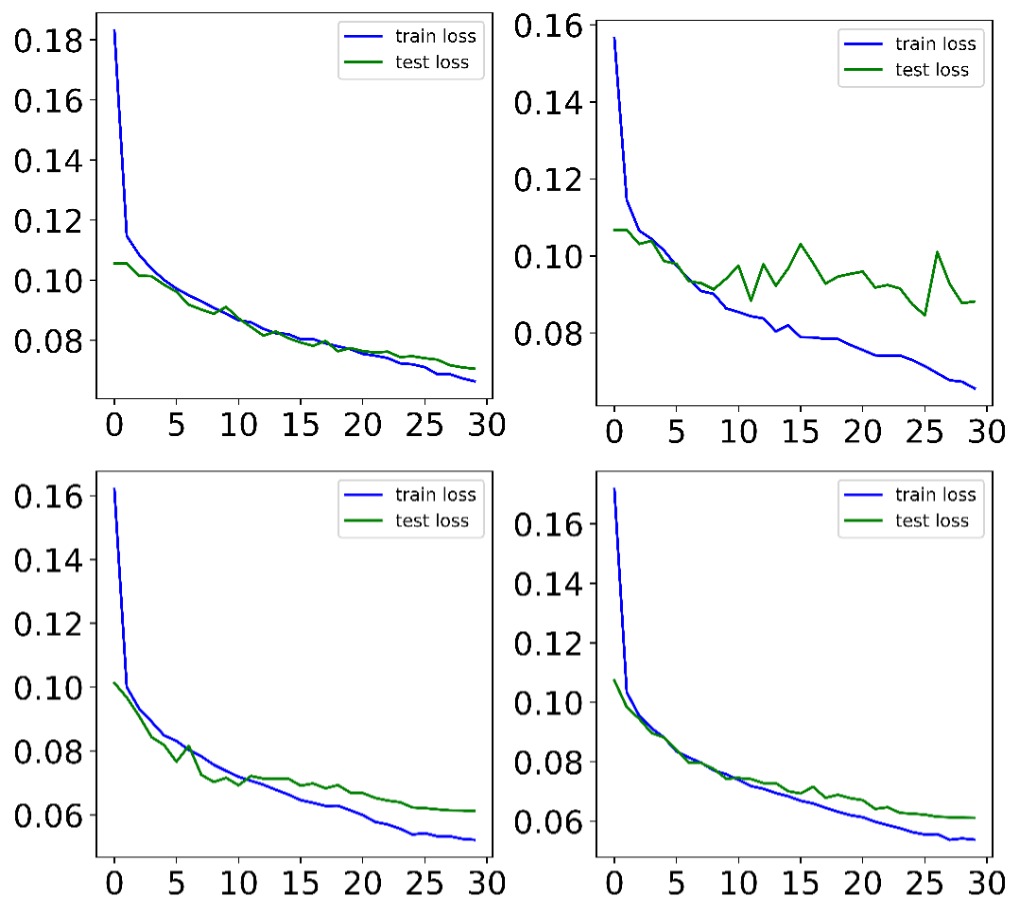}
\end{center}
\vskip-3ex
\caption{A visualization of the loss changing during training and testing on the nuScenes dataset~\cite{nuscenes}. The upper left figure indicates the pillar model and the upper right figure indicates the pillar model under noise disturbance. The bottom left figure indicates our MASS model and the bottom right figure indicates MASS under noise disturbance.}
\label{fig:trainval}
\end{figure}
Comparing the experimental results shown in Table~\ref{tab:experiments_nusc}, under \emph{Noise Ablation} mode and \emph{Dense Train Dense Eval} mode, through addition of the noise under the control condition \emph{SNR=10}, the performance of the model leveraging only pillar feature has a decrease of $6.8\%$, while the performance of MASS has a decrease of $0.6\%$, demonstrating the efficacy of MA against noise. The visualization of the loss changes during training and testing are depicted in Fig.~\ref{fig:trainval} where the upper row denotes the pillar model, while the bottom row denotes the MASS model. The second column denotes training and testing under noise disturbance. According to Fig.~\ref{fig:trainval}, it indicates that the overfitting problem didn't occur in MASS. Comparing the performance of the MASS model and the pillar model under the noise disturbance, MASS shows better performance than the pillar model, and the pillar model shows large fluctuations in testing.
The ablation result of the order of these three attentions is shown in Table~\ref{tab:experiments_nusc} (see \emph{Order Ablation} mode). First, through the comparison among the attention orders PLG, LPG, and LGP, placing pillar attention at the end indicates a better performance. Comparing GLP and LGP, the performance of the model which places Graph attention at the middle shows better performance. This result shows that if we follow the \emph{local-global-local} order and place the pillar attention at the end, the performance of MASS is better following the aforementioned analyses regarding the whole model structure.
\subsection{Cross-Task Analysis of MA for 3D Object Detection}
Our next area of investigation is the cross-task generalization of the proposed MA mechanism.
The prediction results of \emph{pedestrian} and \emph{car}, the most important classes of urban scene, are illustrated.
The first experiment is based on PointPillars~\cite{pointpillars}, which is selected as the baseline for numerical comparison. 
Through the comparison results shown in Table~\ref{tab:3DDetection}, the pillar attention has introduced a performance improvement for \emph{pedestrian} detection in 3D@mAP on the moderate difficulty level.
The results in all the evaluation metrics of \emph{car} have been improved by this attention.
Evidently, \emph{pedestrian} is more difficult to detect due to its small spatial size and also pillar-based method generates pseudo image in the top view, which makes this problem even harder to solve, since \emph{pedestrian} only takes up several pixels on the top-view image. Therefore, to achieve performance improvement of \emph{pedestrian} detection is more difficult than that of \emph{car}.
3D object detection scores on the moderate level can be leveraged to determine the model efficacy, since the sample number is enough while remaining a certain difficulty.
\begin{table}[!t]
\centering
\caption{Quantitative results on the KITTI 3D detection evaluation dataset~\cite{kitti}, where \emph{P} indicates pillar attention, \emph{L} indicates DR LSTM attention, and \emph{G} indicates graph attention.}
\label{tab:3DDetection}
\scalebox{0.9}{\begin{tabular}{lcccccc} 
\toprule
\multicolumn{1}{c}{\multirow{2}{*}{Method}} & \multicolumn{3}{c}{3D@mAP} & \multicolumn{3}{c}{BEV@mAP} \\
\multicolumn{1}{c}{} & Easy & \multicolumn{1}{l}{Mod.} & Hard & \multicolumn{1}{l}{Easy} & Mod. & \multicolumn{1}{l}{Hard} \\ 
\midrule
\multicolumn{7}{c}{Pedestrian} \\ 
\midrule
Pillar & 69.26 & 62.40 & 58.06 & 74.07 & 69.83 & 64.37 \\
Pillar + P & 68.00 & 63.20 & 57.38 & 73.11 & 68.34 & 62.68 \\
Pillar + LP & 70.03 & 64.76 & 59.81 & 74.52 & 69.89 & 64.92 \\
Pillar + LGP & \multicolumn{1}{l}{\textbf{71.39}} & \textbf{65.80} & \multicolumn{1}{l}{\textbf{60.11}} & \textbf{77.48} & \textbf{71.23} & \textbf{65.39} \\ 
\midrule
\multicolumn{7}{c}{Car} \\ 
\midrule
Pillar & 86.09 & 74.10 & 69.12 & 89.78 & 86.34 & 82.08 \\
Pillar + P & 86.36 & 76.73 & 70.20 & \textbf{90.09} & \textbf{87.22} & \textbf{85.57} \\
Pillar + LP & 86.59 & 76.13 & 70.40 & 89.90 & 87.03 & 84.94 \\
Pillar + LGP & \textbf{87.47} & \textbf{77.03} & \textbf{73.25} & 89.94 & 87.09 & 84.80 \\
\bottomrule
\end{tabular}}
\vskip-3ex
\end{table}

We observe that the improvement performance by the pillar attention mechanism of $0.80\%$ for \emph{pedestrian} on the moderate level for 3D@mAP, when compared to the raw PointPillars~\cite{pointpillars} indicated by \emph{Pillar}.
Besides, there is also a gain of $2.63\%$ on moderate 3D@mAP for \emph{car}, indicating that the attention generated through point-wise and channel-wise aggregations inside a pillar is effective for high-level discriminative feature representations. 
Next, we validate PointPillars equipped with the pillar attention and DR LSTM attention. All evaluation metrics both for
3D@mAP and BEV@mAP of these two classes are consistently improved through this enhancement. 
It turns out that DR LSTM attention is efficient for producing attention values guiding the model to focus on the significant pillars for 3D object detection, as it takes in consideration of aggregated local information. 
The 3D@mAP score has a $2.36\%$ improvement on \emph{pedestrian} and a $2.03\%$ improvement on \emph{car} on the moderate difficulty level.

Finally, the last experiment concerns combining PointPillars with MA, meaning that all the attention-based building blocks are leveraged: the pillar attention, DR LSTM attention, and key-node based feature-steered graph attention.
MA leads to a $3.40\%$ performance gain for \emph{pedestrian} on the moderate level 3D@mAP and a $2.93\%$ performance improvement for \emph{car}, which is the best model during experiments.
Since DR LSTM attention preserves locality, global attention generation mechanism such as the graph attention proposed by our work is able to aggregate more important cues from key nodes generated through FPS on the high-level feature space and propagate these information to the others. 
Overall, the experiment results demonstrate the effectiveness of our MA model for generalizing to 3D detection.
\subsection{Cross-Task Approaches Analyses and Comparisons}
In the following, we compare our MASS approach with GndNet~\cite{gndnet}, RangeNet++~\cite{rangenet++} and PolarNet~\cite{polarnet} which are focusing on different domain outputs for semantic segmentation while using the same dataset SemanticKITTI~\cite{semantic_kitti} and sparse LiDAR data as input. We conduct the analyses according to Table \ref{tab:experiments_cross_field}. First, we conduct the comparison between different approaches based on the output results. For our top-view based approach, it contains less distortions and conserves affine invariance compared with the panoramic-view based approach, RangeNet++~\cite{rangenet++}, which indicates that MASS has great potentiality to make the sub-tasks of automated vehicles such as route planning easier. At the same time, compared with PolarNet~\cite{pointnet} which outputs sparse top-view semantic segmentation, our proposed approach gives more information on the unknown grid cell region which can give more reference information for the automated vehicle to make decisions for the blind zone of LiDAR. Compared with GndNet~\cite{gndnet} which predicts point-wise semantic segmentation category for each 3D LiDAR point, the top-view dense semantic segmentation map encodes higher-level semantic meanings especially on the region where laser ray doesn't travel than sparse 3D point-wise semantic segmentation since the predicted top-view map can be used in several automated vehicle's sub-task applications such as decision making and it has indicated the boundary of each class, while sparse point-wise semantic segmentation prediction needs more postprocessing procedures. Second, considering the performance and inference time, our approach has a relatively higher performance than the other approaches which predict 2D semantic segmentation map while has a relatively decent inference speed. GndNet~\cite{gndnet} has a better score and smaller inference time, but the task difference is huge between GndNet predicting 3D point-wise sparse semantic segmentation and other approaches predicting 2D semantic segmentation including our proposed method. Overall, MASS has great competitiveness even compared with cross-task approaches.
\begin{table}[!t]
\centering
\caption{A comparison between MASS and several cross-field 3D point cloud based semantic segmentation approaches, where \emph{O.Perspective} indicates the point of view of output results.}
\label{tab:experiments_cross_field}
\scalebox{0.9}{
\begin{tabular}{l|llll} 
\toprule
Approach & \multicolumn{1}{c}{MASS} & \multicolumn{1}{c}{GndNet~\cite{gndnet}} & \multicolumn{1}{c}{RangeNet++~\cite{rangenet++}} & \multicolumn{1}{c}{PolarNet~\cite{polarnet}} \\ 
\midrule
O.Perspective & TopView & 3D Space & \textcolor[rgb]{0.125,0.129,0.141}{PanoramicV}iew & TopView \\
Dense/Sparse & Dense & Sparse & Dense & Sparse \\
Inference time & 74ms & 18ms & 83ms & 62ms \\
Optimizer & Adam & SGD & SGD & Adam \\
Score (mIoU) & 58.80 & 84.01 & 52.2 & 54.3 \\
\bottomrule
\end{tabular}}
\vskip-3ex
\end{table}
\subsection{Inference Time}
The inference time of our model without MA and occupancy feature is measured on an NVIDIA GTX2080Ti GPU processor,
achieving a total runtime of $58ms$ per input for dense top-view semantic segmentation on SemanticKITTI. MA doubles the inference runtime compared with the model without MA and occupancy feature. For the model with occupancy feature and without MA, additional $16ms$ are required for the preprocessing and model inference. 
Thereby, MASS has achieved a near real-time speed suitable for transportation applications.

\subsection{Ablation Study on Data Augmentation}
The diversity of training data is crucial for yielding a robust segmentation model in real traffic scenes~\cite{pass}.
We therefore benchmark different data augmentation approaches in our system that are studied and verified through ablation experiments.
According to the results shown in Table~\ref{tab:experiments_da}, the model only with pillar feature and without any data augmentation is chosen as the baseline since it has the fastest inference speed in the \emph{Sparse Eval} mode.
Through observation, \emph{random scale} brings a $0.6\%$ mIoU improvement, while \emph{random flip} and \emph{random rotation} significantly improve mIoU by $4.6\%$, which helps to yield robust models for dense top-view semantic segmentation.
The \emph{random translation} does not contribute to any performance improvement since it moves the position of ego car of each LiDAR frame, and therefore it is not recommended.
Overall, with these data augmentation operations, we have further improved the generalization capacity of the proposed model for real-world $360^\circ$ surrounding understanding.
\begin{table}[!t]
\centering
\caption{Ablation study for data augmentation techniques on the SemanticKITTI dataset~\cite{semantic_kitti}.}
\label{tab:experiments_da}
\scalebox{0.9}{\begin{tabular}{cccccc}
\toprule
Baseline & Flip & Rotate & Scale & Translate & mIoU [\%] \\
\midrule
\checkmark & & & & & 50.4 \\
\checkmark & \checkmark & & & & 53.0 \\
\checkmark & \checkmark & \checkmark & & & 55.0 \\
\checkmark & \checkmark & \checkmark & \checkmark & & 55.6 \\
\checkmark & \checkmark & \checkmark & \checkmark & \checkmark & 55.1 \\
\bottomrule
\end{tabular}}
\vskip-2ex
\end{table}
\section{Conclusion}
In this work, we established a novel Multi-Attentional Semantic Segmentation (MASS) framework for dense surrounding understanding of road-driving scenes.
A pillar-based end-to-end approach enhanced with Multi-Attention (MA) mechanism is presented for dense top-view semantic segmentation based on sparse LiDAR data.
Pillar-based representations are learned end-to-end therefore avoiding information bottlenecks compared with handcrafted features leveraged in grid maps based approach~\cite{bieder2020exploiting}.
Extensive model ablations consistently demonstrate the effectiveness of MA on dense top-view semantic segmentation and 3D object detection.
Our quantitative experiments highlight the quality of our model predictions, surpassing existing  state-of-the-art methods.

In the future, we aim to build on the top-view semantic segmentation approach and investigate cross-dimensional semantic mapping for various automated transportation applications.
From the algorithmic perspective, we intend to extend and study our framework with unsupervised domain adaptation and dense contrastive learning strategies for uncertainty-aware driver behavior and holistic scene understanding. We also intend to reformulate the work procedure of PFN and densify the annotation for moving objects to reduce information loss generated through pillarization.

\bibliographystyle{IEEEtran}
\bibliography{bib}

\end{document}